\documentclass[11pt]{article}

\usepackage[preprint]{acl}

\usepackage{times}
\usepackage{latexsym}

\usepackage{hyperref}
\usepackage{url}
\usepackage{tabularx}
\usepackage{booktabs}
\usepackage{multirow}
\usepackage{multicol}
\usepackage{soul}
\usepackage{graphicx}
\usepackage{xcolor,colortbl}
\definecolor{lightgray}{gray}{0.9}
\usepackage{bbding}
\usepackage{pifont}
\usepackage{diagbox}
\usepackage{longtable}
\usepackage{colortbl}
\usepackage{rotating}
\usepackage{wrapfig}
\usepackage[normalem]{ulem}
\usepackage[most]{tcolorbox}
\usepackage{subcaption}

\usepackage[T1]{fontenc}

\usepackage[utf8]{inputenc}

\usepackage{microtype}

\usepackage{inconsolata}

\usepackage{graphicx}

\title{Natural Context Drift Undermines the Natural\\ Language Understanding of Large Language Models}

\author{Yulong Wu\textsuperscript{1}, Viktor Schlegel\textsuperscript{1, 2} and Riza Batista-Navarro\textsuperscript{1} \\
  \textsuperscript{1} Department of Computer Science, University of Manchester, United Kingdom \\
  \textsuperscript{2} Imperial College London, Imperial Global Singapore \\
  \texttt{\{yulong.wu, riza.batista\}@manchester.ac.uk} \\
  \texttt{v.schlegel@imperial.ac.uk}\\}

\begin{document}
\maketitle
\begin{abstract}

How does the natural evolution of context paragraphs affect question answering in generative Large Language Models (LLMs)? To investigate this, we propose a framework for curating naturally evolved, human-edited variants of reading passages from contemporary QA benchmarks and for analyzing LLM performance across a range of semantic similarity scores, which quantify how closely each variant aligns with content seen during pretraining. Using this framework, we evaluate six QA datasets and eight LLMs with publicly available training data. Our experiments reveal that LLM performance declines as reading passages naturally diverge from the versions encountered during pretraining—even when the question and all necessary information remains present at inference time. For instance, average model accuracy on \textsc{BoolQ} drops by over $30$\% from the highest to lowest similarity bins, with slopes exceeding $70$ across several LLMs. These findings suggest that natural text evolution poses a significant challenge to the language understanding capabilities of LLMs.

\end{abstract}

\section{Introduction}
\label{sec:Introduction}

Large Language Models (LLMs), pre-trained on massive web-scale corpora, have proven effective at Question Answering (QA) over text passages~\citep{openai2024gpt4technicalreport, deepseekai2025deepseekv3technicalreport, deepseekai2025deepseekr1incentivizingreasoningcapability, yang2025qwen3technicalreport, olmo20252olmo2furious}, a task that has long been established as a testbed for evaluating natural language understanding~\citep{chen2018neural}. Nonetheless, concerns remain regarding their genuine reading comprehension abilities and generalization, as revealed by research efforts on robustness evaluation~\citep{wu-etal-2023-machine, levy-etal-2023-guiding}, benchmark contamination impact analysis~\citep{palavalli-etal-2024-taxonomy, li-etal-2024-open-source}, and others.

Differentiating from previous work, this paper offers a new perspective on understanding the limitations of generative LLMs by asking: what happens when reading paragraphs continue to evolve and diverge from their appearance during pretraining? This scenario is common in real-world applications, where test data naturally changes over time due to ongoing human edits, content updates, or shifts in context, (e.g., Wikipedia articles~\citep{yang-etal-2017-identifying-semantic}), and therefore requires genuine language understanding from LLMs. To the best of our knowledge, however, no prior work has systematically investigated this phenomenon in QA.

To address this gap, we propose a framework to analyse how LLMs performance changes as the reading paragraph semantically diverges from the content of its source in the model’s training corpora. Among various examples of evolving text corpora, we focus on Wikipedia, as it serves as a primary source for reading passages in widely used QA benchmarks~\citep{wang2022modernquestionansweringdatasets}, is commonly included in LLM training~\citep{zhao2025surveylargelanguagemodels}, and most importantly, the evolution of text is clearly documented via revision histories. This enables us to curate human-edited variants of passages that reflect natural text evolution over time. Our approach adopts a gradual perspective by computing a continuous semantic similarity score at the paragraph level and correlating it with LLM's QA accuracy.

Within the developed framework, we empirically evaluate six QA benchmark datasets and eight LLMs with fully open-source training data. Our study finds that, across models with different training corpora and architectural configurations, as context paragraphs naturally evolve and become semantically distant from the Wikipedia content sharing the same article title seen during pretraining, the reading comprehension performance of LLMs generally deteriorates. In contrast, human annotators are less affected by such semantic drift and maintain relatively stable accuracy regardless of passage similarity, suggesting that the observed performance drop is specific to LLMs and not due to deficiencies in the edited passages themselves.

\begin{figure*}
    \centering
    \includegraphics[width=\textwidth]{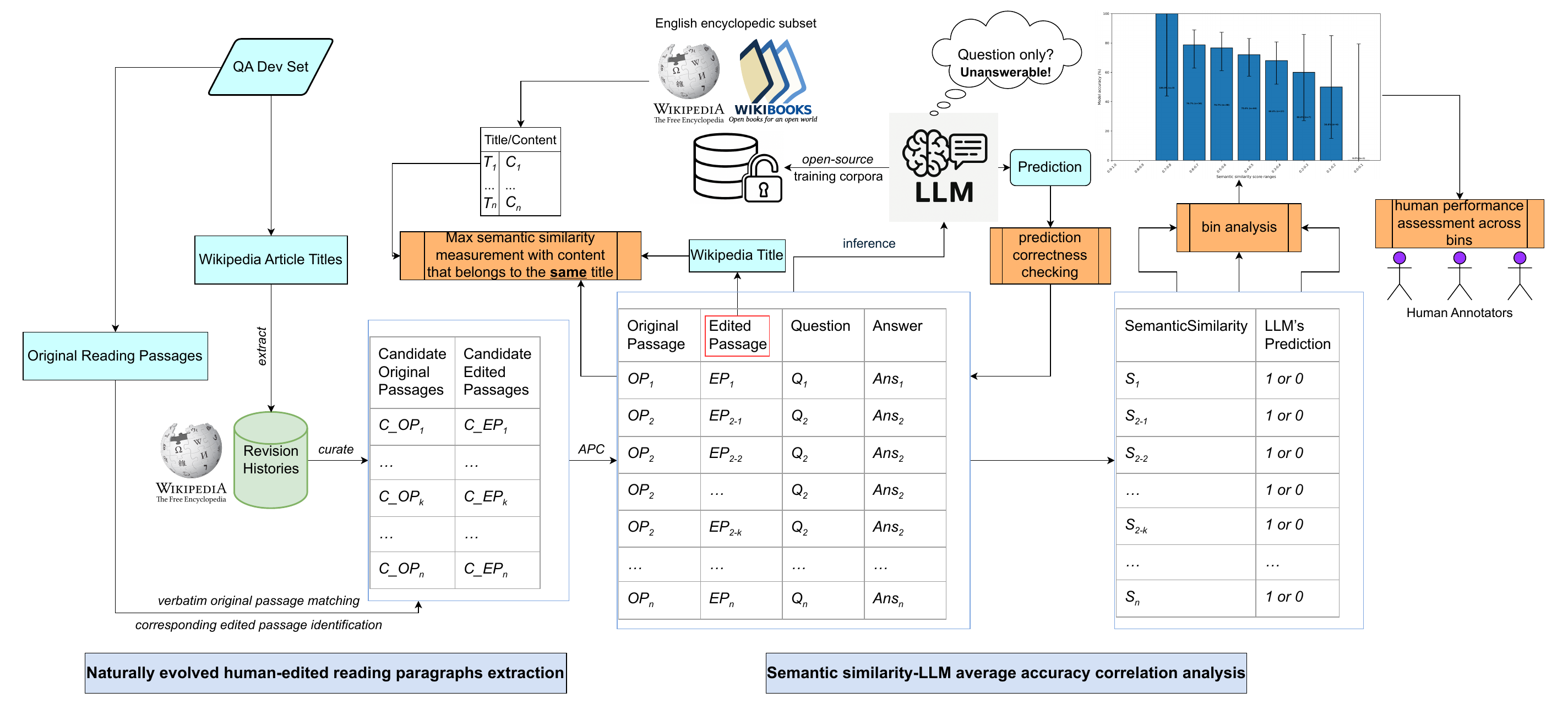}
    \caption{An overview of the analysis framework. Module~\textit{Naturally evolved human-edited reading paragraphs extraction} is adapted from \citep{wu2025payattentionrealworld} with minor modifications. APC: Answers Preserving Checking.}
    \label{fig:methodology}
\end{figure*}

\section{Methodology}
\label{sec:Methodology}

In our framework (Figure~\ref{fig:methodology}), we extract revision histories of paragraphs from QA benchmarks, order them by semantic similarity to the version that appears in an LLM's training corpus, and correlate the LLMs' answer accuracy on those passages to the similarity thus obtained. The framework consists of two modules, described in detail below.

\paragraph{Naturally evolved human-edited reading paragraphs extraction.} To obtain edited versions of original reading paragraphs from contemporary QA benchmark datasets that genuinely reflect real-world scenarios, we adopt the natural perturbation pipeline proposed by~\citet{wu2025payattentionrealworld}, with two slight modifications: 1) we remove the constraint of retaining only candidate passage pairs where both paragraphs exceed 500 characters, allowing broader dataset applicability and preservation of diverse editing patterns; and 2) for the matched
original passages with multiple occurrences, we retain all edited versions for each (see passage \textit{OP\textsubscript{2}} in Figure~\ref{fig:methodology} as an example) to support subsequent correlation analysis. Appendix~\ref{sec:Answer Preservation Check and Data Statistics} provides details on Answers Preserving Checking and data statistics.

\paragraph{Semantic similarity-LLM average accuracy correlation analysis.} For each naturally evolved, human-edited reading paragraph and its associated question, we generate predictions using an LLM and label them as 1 (correct) or 0 (incorrect), based on a selected evaluation metric. We also collect predictions using the question alone to test whether the LLM already possesses parametric knowledge of the answer. Instances in which the LLM answers correctly without access to the passage are excluded, as they may call into question the paragraph’s contribution to the answer \citep{glockner2025neoqaevidencebasedquestionanswering}~\footnote{Appendix~\ref{sec:Percentage of Instances where LLMs Succeed on Context-Free Question Answering} shows that, within each generated dataset in Module~\textit{Naturally evolved human-edited reading paragraphs extraction}, the percentage of instances in which \texttt{OLMo} LLMs succeed on context-free QA, and are thus filtered out.}. Meanwhile, we extract English Wikipedia content from the LLM’s training corpora that shares the same article title as the edited passage and compute the semantic similarity between them. The maximum similarity score is used as a proxy for how closely the passage resembles the training data. We then group the similarity scores into ten bins, compute the average LLM accuracy within each bin, and plot accuracy trends from highest to lowest similarity. Uncertainty for the result in each bin is estimated using the Wilson score interval with 95\% confidence ($z$ = 1.96) \citep{Wilson01061927}. Finally, to validate the observed trend, we assess human performance across the same bins.

\section{Experiments Setup}
\label{sec:Experiments Setup}

\begin{figure*}[htb!]
    \centering
    \begin{subfigure}[b]{0.3\textwidth}
        \includegraphics[width=\textwidth]{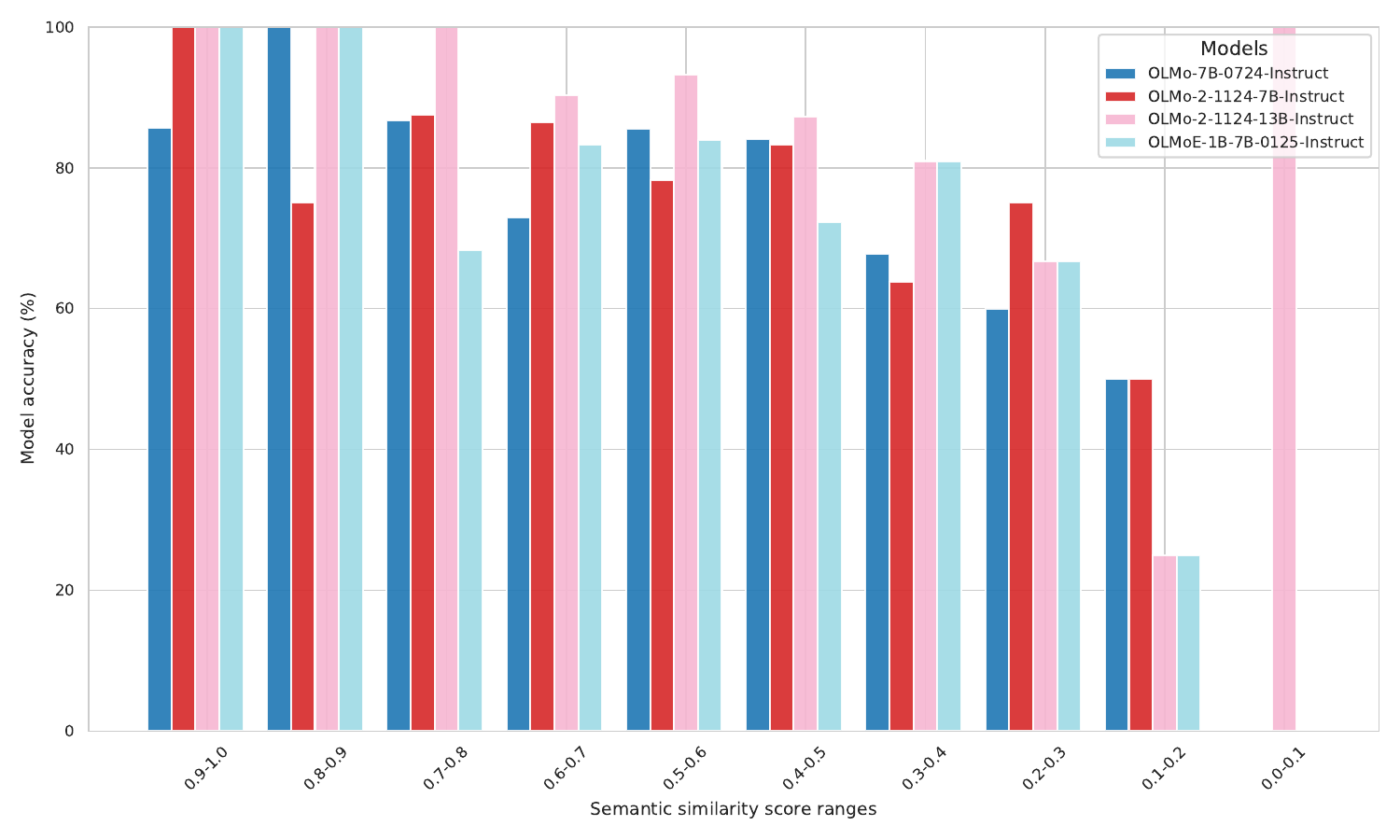}
        \caption{\textsc{SQuAD 1.1}}
    \end{subfigure}
    \hfill
    \begin{subfigure}[b]{0.3\textwidth}
        \includegraphics[width=\textwidth]{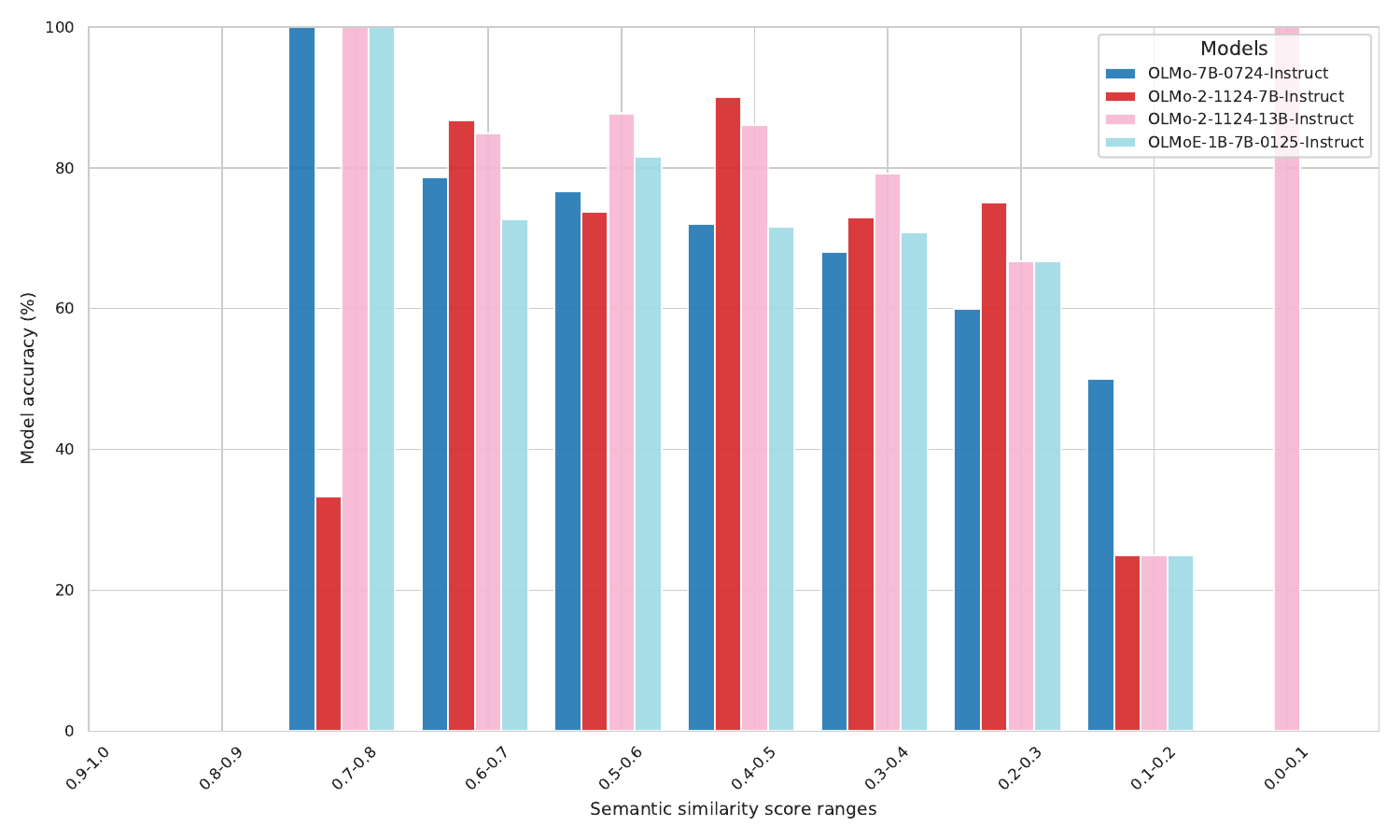}
        \caption{\textsc{SQuAD 2.0}}
    \end{subfigure}
    \hfill
    \begin{subfigure}[b]{0.3\textwidth}
        \includegraphics[width=\textwidth]{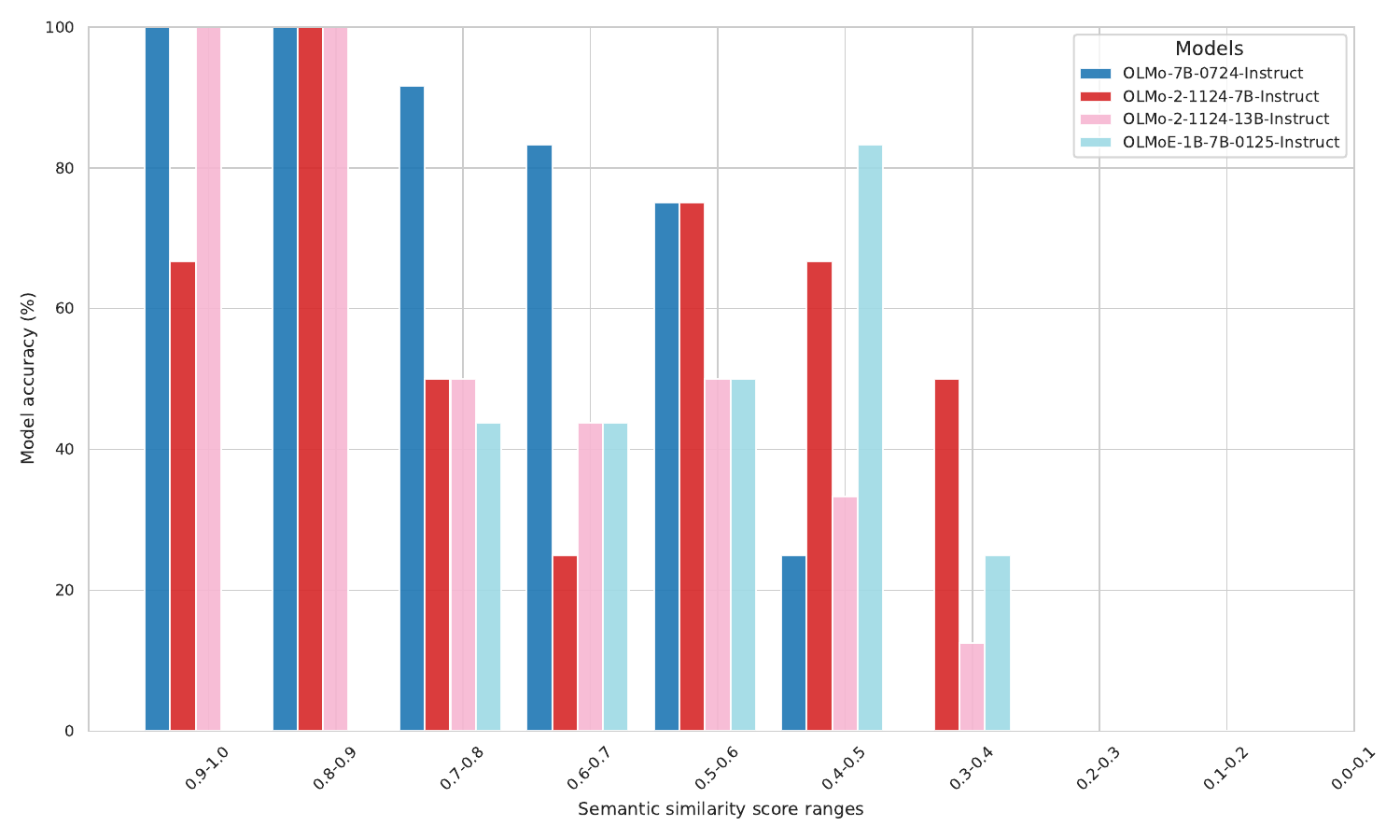}
        \caption{\textsc{D(RoBERTa)}}
    \end{subfigure}

    \vspace{0.5cm}

    \begin{subfigure}[b]{0.3\textwidth}
        \includegraphics[width=\textwidth]{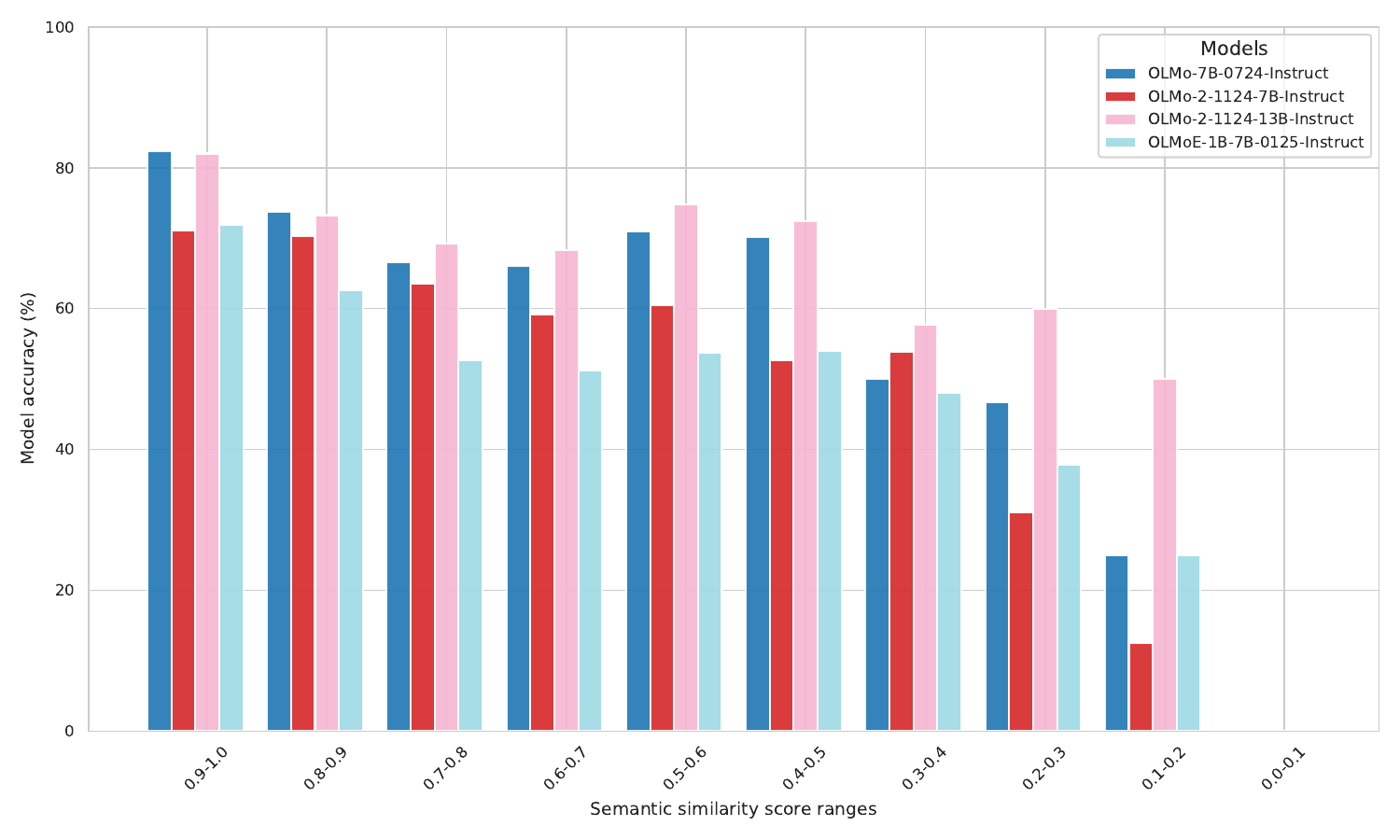}
        \caption{\textsc{BoolQ}}
    \end{subfigure}
    \hfill
    \begin{subfigure}[b]{0.3\textwidth}
        \includegraphics[width=\textwidth]{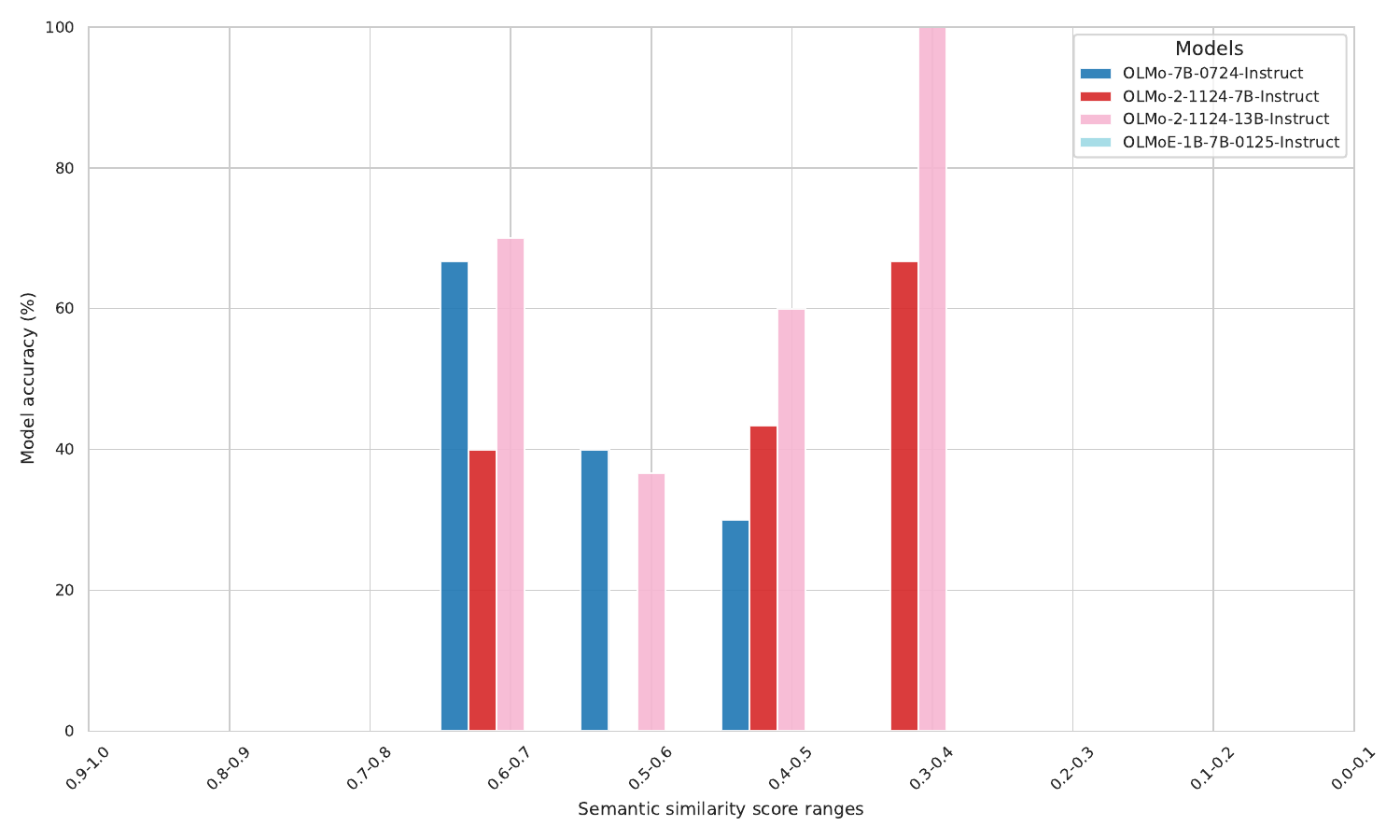}
        \caption{\textsc{WikiWhy}}
    \end{subfigure}
    \hfill
    \begin{subfigure}[b]{0.3\textwidth}
        \includegraphics[width=\textwidth]{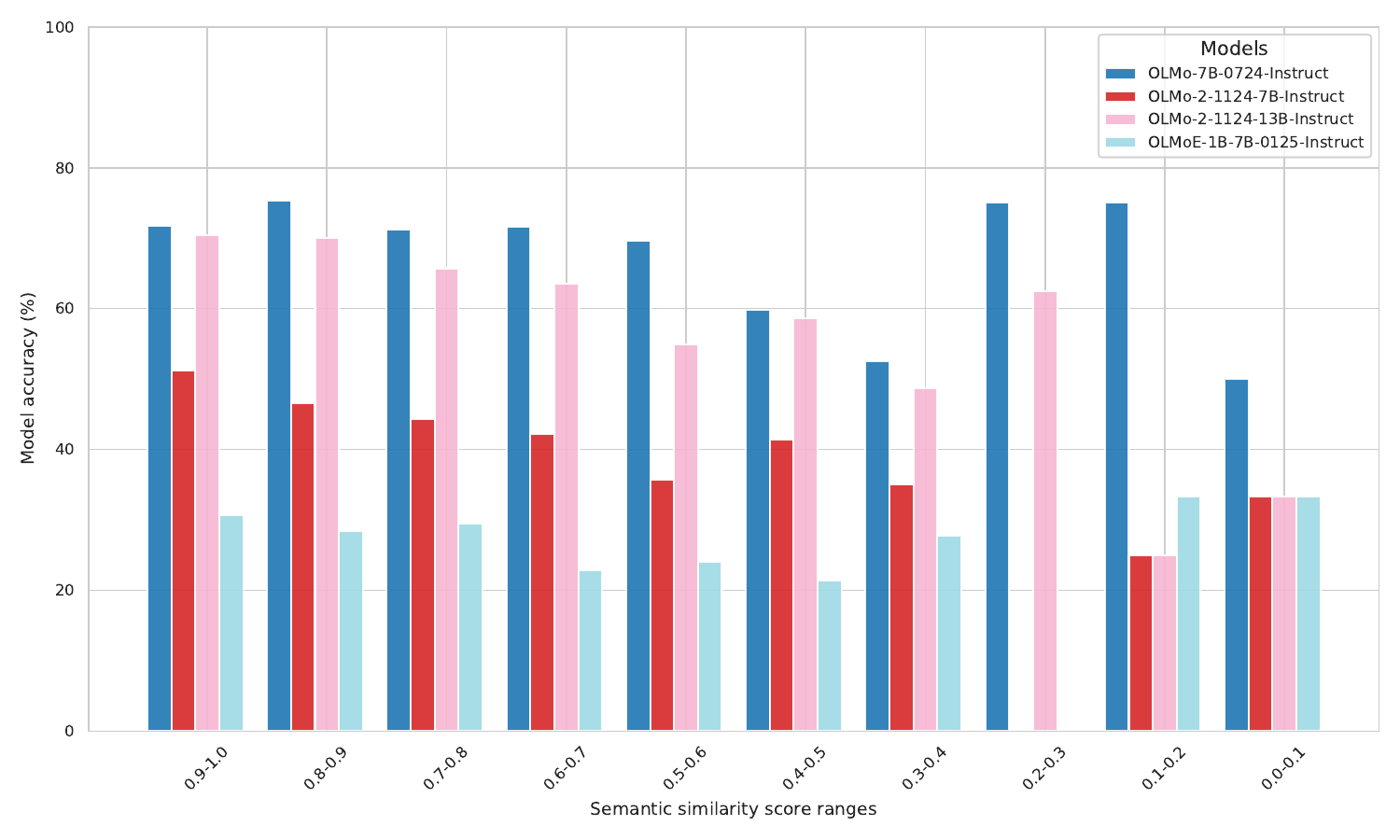}
        \caption{\textsc{HotpotQA}}
    \end{subfigure}

    \caption{Average accuracy of the instruction-finetuned \texttt{OLMo} LLMs across ten semantic similarity bins.}
    \label{fig:olmo_results}
\end{figure*}

\begin{figure*}[htb!]
    \centering
    \begin{subfigure}[b]{0.3\textwidth}
        \includegraphics[width=\textwidth]{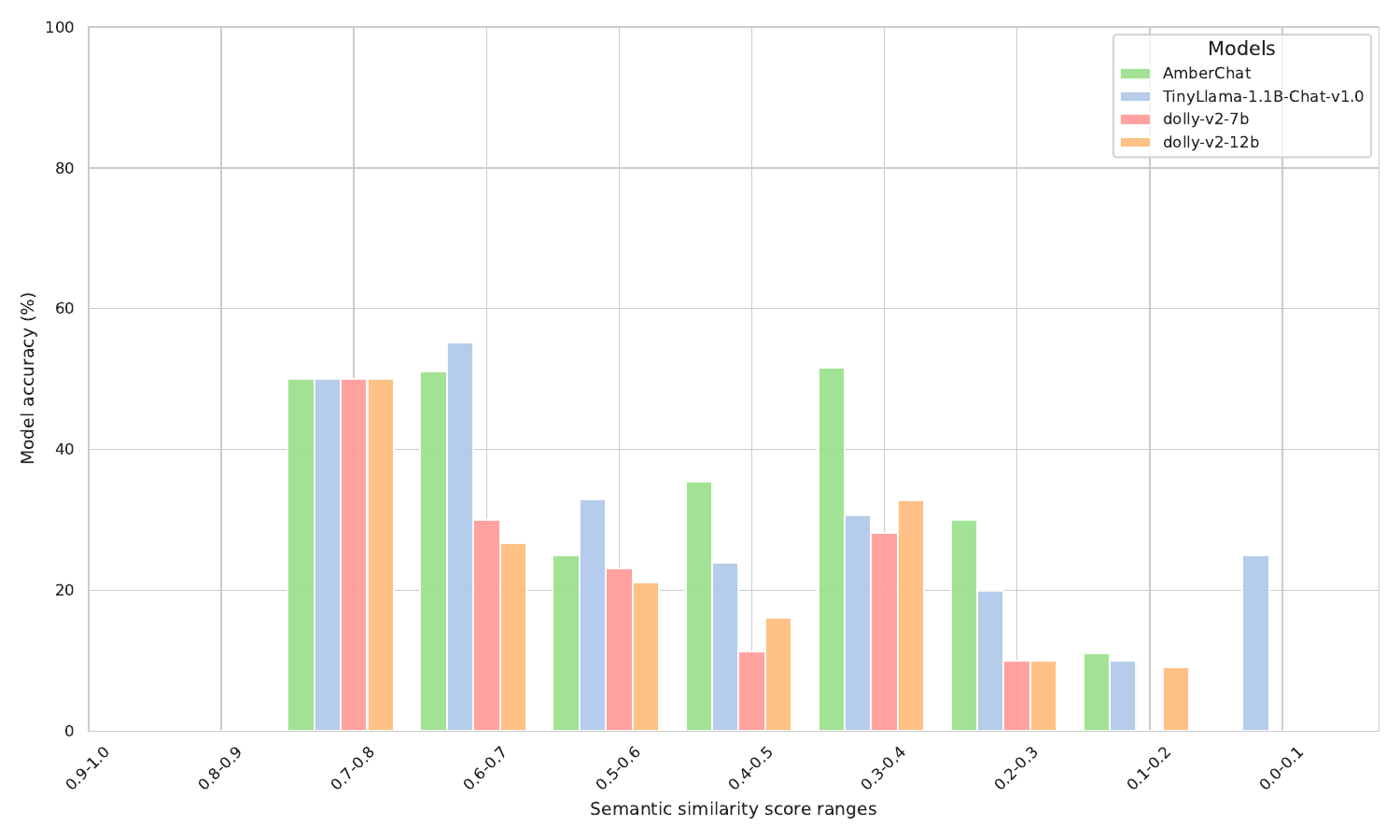}
        \caption{\textsc{SQuAD 2.0}}
    \end{subfigure}
    \hfill
    \begin{subfigure}[b]{0.3\textwidth}
        \includegraphics[width=\textwidth]{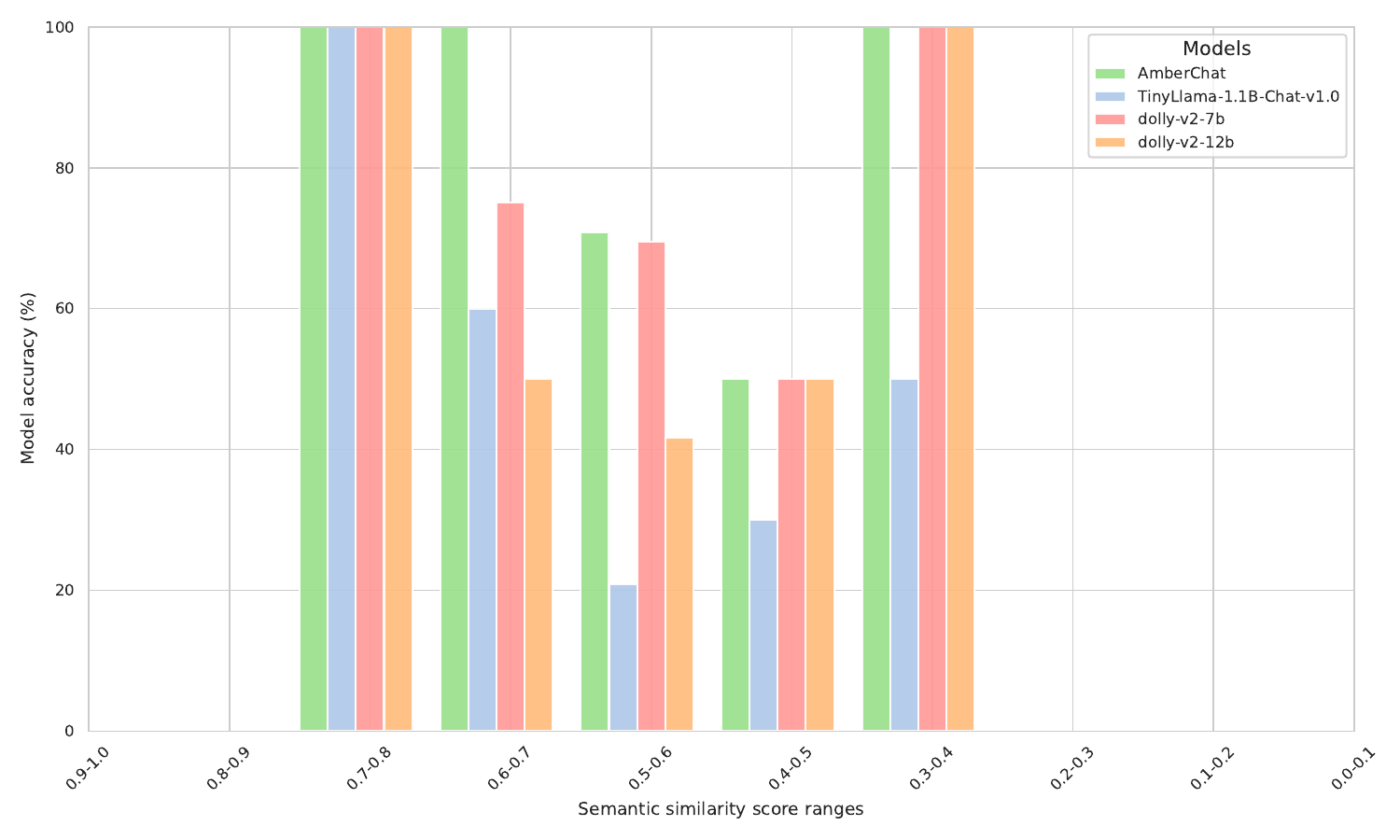}
        \caption{\textsc{WikiWhy}}
    \end{subfigure}
    \hfill
    \begin{subfigure}[b]{0.3\textwidth}
        \includegraphics[width=\textwidth]{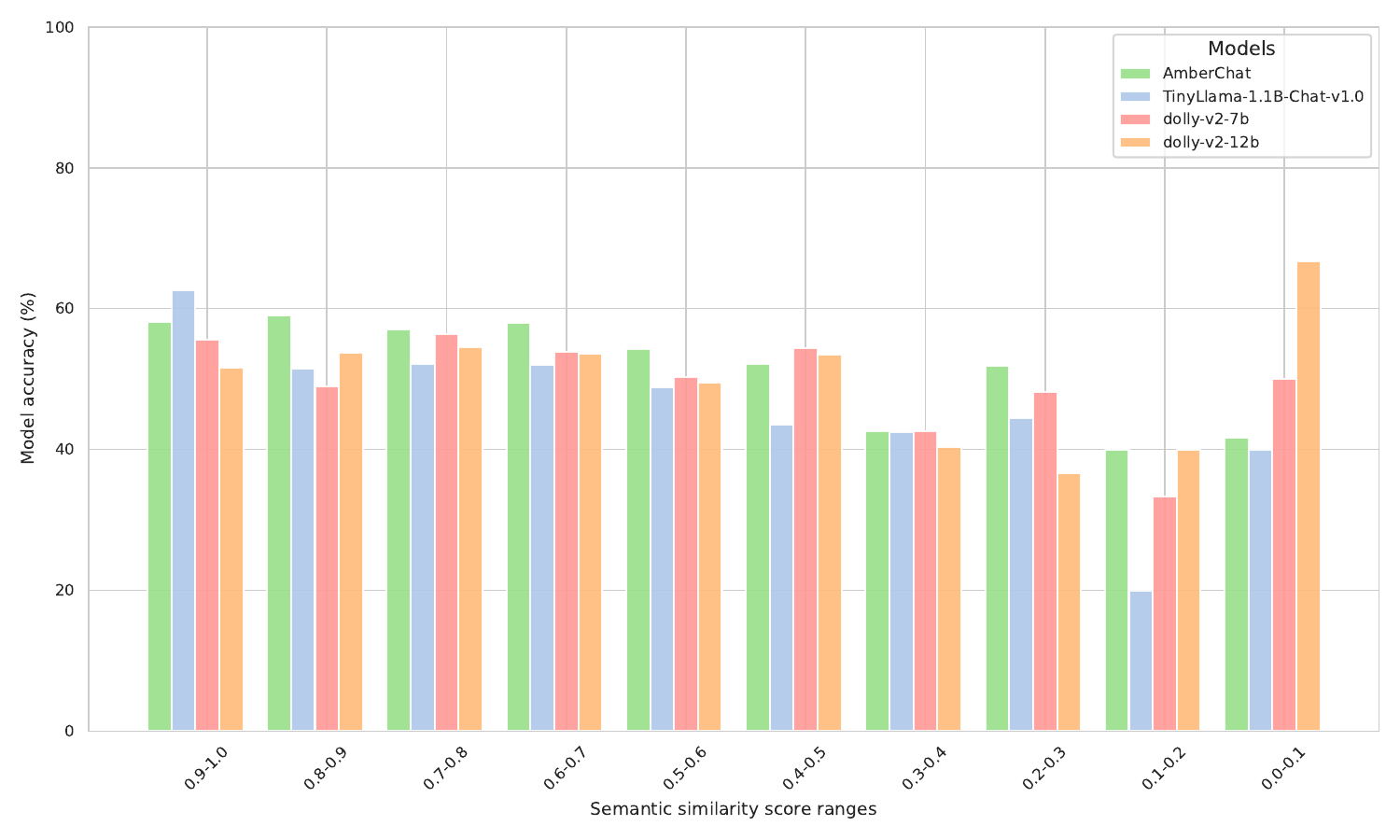}
        \caption{\textsc{HotpotQA}}
    \end{subfigure}
    \caption{Average accuracy of other instruction-finetuned LLMs across ten semantic similarity bins.}
    \label{fig:other_results}
\end{figure*}

Broadly, we address the following question: \textit{How well do LLMs perform on QA as the reading paragraphs naturally evolve from the versions present in their training corpus?} To this end, we select QA benchmarks that feature context paragraphs from Wikipedia, whose edit histories allow us to trace naturally evolved versions of each paragraph, and evaluate LLMs with an open-source training corpus, as detailed below.

\textbf{Datasets:} We use the development set of six English QA datasets spanning extractive, yes/no, abstractive, cause-effect reasoning and multi-hop reasoning challenge: \textsc{SQuAD 1.1} \citep{rajpurkar-etal-2016-squad}, \textsc{SQuAD 2.0} \citep{rajpurkar-etal-2018-know}, \textsc{AdversarialQA - D(RoBERTa)} \citep{bartolo-etal-2020-beat}, \textsc{BoolQ} \citep{clark-etal-2019-boolq}, \textsc{WikiWhy} \citep{ho2023wikiwhy} (version 1.2) and \textsc{HotpotQA} \citep{yang-etal-2018-hotpotqa} (in the ``distractor'' setting). For \textsc{HotpotQA}, which includes paragraphs from multiple Wikipedia articles within a single context, we retain only the edits applied to the gold passages, i.e., those containing supporting facts that determine the answer, and disregard other distractors.

\textbf{LLMs:} We evaluate eight transparent instruction-tuned LLMs across six model families, all of which include Wikipedia as part of their publicly available training data: \texttt{\textbf{OLMo}} (\texttt{OLMo-7B-0724-Instruct-hf}) \citep{groeneveld-etal-2024-olmo}, \texttt{\textbf{OLMo 2}} (\texttt{OLMo-2-1124-7B-Instruct} and \texttt{OLMo-2-1124-13B-Instruct}) \citep{olmo20252olmo2furious}, \texttt{\textbf{OLMoE}} (\texttt{OLMoE-1B-7B-0125-Instruct}) \citep{muennighoff2025olmoeopenmixtureofexpertslanguage}, LLM360's \texttt{\textbf{AmberChat}} \citep{liu2024llm}, \texttt{\textbf{TinyLlama}} (\texttt{TinyLlama-1.1B-Chat-v1.0}) \citep{zhang2024tinyllamaopensourcesmalllanguage}, Databricks' \texttt{\textbf{Dolly}} (\texttt{dolly-v2-7b} and \texttt{dolly-v2-12b}) \citep{DatabricksBlog2023DollyV2}. To isolate the potential impact of instruction tuning, we also conduct experiments on the base (non-aligned) versions of the same models. All LLMs are tested in a zero-shot setting, and the inference prompts are provided in Appendix~\ref{sec:qa_prompts}. Model experimentation is carried out using the HuggingFace's~\textit{Transformers} library \citep{wolf-etal-2020-transformers}, the vLLM \citep{10.1145/3600006.3613165}, and two 80GB NVIDIA A100 GPUs. LLM prediction correctness is determined using Inclusion Match (IM), which considers a prediction correct if it includes any ground truth answer \citep{levy-etal-2023-guiding, bhuiya-etal-2024-seemingly}. For \textsc{WikiWhy}, due to its free-form format, correctness is based on semantic similarity (using the~\texttt{all-MiniLM-L6-v2} model) between the LLM’s response and the ground truth answer; predictions scoring above $0.6$ are considered correct.

\textbf{Semantic Similarity Measure:} We measure semantic-level textual similarity between the edited Wikipedia paragraph and the versions found in the Wikipedia subset of each LLM’s training corpus--sourced from \textsc{Dolma v1.7}\footnote{\url{https://huggingface.co/datasets/allenai/dolma}} \citep{soldaini-etal-2024-dolma} for \texttt{OLMo}, \textsc{RedPajama v1}\footnote{\url{https://huggingface.co/datasets/togethercomputer/RedPajama-Data-1T}} \citep{together2023redpajama} for \texttt{AmberChat}, \textsc{SlimPajama}\footnote{\url{https://huggingface.co/datasets/cerebras/SlimPajama-627B}} \citep{cerebras2023slimpajama} for \texttt{TinyLlama} and \textsc{Pile}\footnote{As \textsc{Pile} is no longer officially hosted or distributed, we use a publicly accessible replication of its original Wikipedia component: \url{https://github.com/noanabeshima/wikipedia-downloader}.} \citep{gao2020pile800gbdatasetdiverse, biderman2022datasheetpile} for \texttt{Dolly}, using a Sentence Transformers model \texttt{all-MiniLM-L6-v2} \citep{reimers-gurevych-2019-sentence}. In addition, we employ three alternative embedding models for the similarity measure to enhance the generalizability of our study: \texttt{sentence-t5-base} \citep{ni-etal-2022-sentence}, \texttt{all-mpnet-base-v2} and \texttt{bge-small-en-v1.5} \citep{10.1145/3626772.3657878}. In measuring semantic similarity for~\textsc{HotpotQA}, we consider only the paragraphs in the edited context whose corresponding Wikipedia article titles have matching content in the Wikipedia subset of an LLM's training corpus.

\section{Results and Discussion}
\label{sec:Results and Discussion}

\textit{\textbf{As a general trend, the QA performance of instruction-finetuned \texttt{OLMo} LLMs deteriorates as the reading paragraph semantically deviates from the training corpus.}} This is clearly visualized in Figure~\ref{fig:olmo_results}, where, across the evaluated benchmark datasets, model average accuracy generally declines as the reading paragraphs evolve and exhibit lower semantic similarity to the versions found in the Wikipedia subset of the training corpus. For example, on \textsc{BoolQ}, the accuracy of \texttt{OLMo-2-1124-7B-Instruct} declines sharply from 71.1\% in the highest similarity bin 0.9-1.0 to 12.5\% in the lowest 0.1-0.2. Comparable drops are also observed, for instance, for \texttt{OLMo-7B-0724-Instruct} on \textsc{SQuAD 2.0} and \textsc{WikiWhy}, and for \texttt{OLMo-2-1124-13B-Instruct} on \textsc{D(RoBERTa)} and \textsc{HotpotQA}, illustrating the broader impact of natural context drift in diverse QA challenges. To further substantiate the observed trend, we perform a slope analysis as shown in Appendix~\ref{sec:Slope Analysis of Accuracy vs. Semantic Similarity} Figure~\ref{fig:slope}, where linear regressions are fitted to each model’s accuracy trajectory across semantic similarity bins. Aggregated across all tasks and models, the mean slope is 65.78 ± 41.55, with a mean Pearson correlation of 0.684 ± 0.291, highlighting a consistent and statistically grounded relationship between semantic divergence and performance degradation. Further, this downward trend is not exclusive to the \texttt{OLMo} family. As shown in Figure~\ref{fig:other_results}, \textit{\textbf{a comparable decline in average performance across decreasing similarity ranges is likewise observed in other instruction-finetuned LLMs, including \texttt{AmberChat}, \texttt{TinyLlama-1.1B-Chat-v1.0} and \texttt{dolly-v2-7b/12b}, despite differences in model size, architecture, training corpora and procedure.}} To further investigate the generalisability of our findings, we conduct an ablation study using alternative embedding models for semantic similarity measurement (Figure~\ref{fig:olmo_results_alternative_metrics} in Appendix~\ref{sec:Consequences of Context Drift Persist Across Other Embedding Models}), evaluating the base pre-trained versions of the instruction-finetuned LLMs (Figure~\ref{fig:olmo_results_Base} and Figure~\ref{fig:other_results_Base} in Appendix~\ref{sec:Context Drift Effects Hold Without Instruction Tuning}) and prompting LLMs to perform chain-of-thought (CoT) reasoning \citep{10.5555/3600270.3602070} when generating responses (Figure~\ref{fig:olmo_results_CoT} and Figure~\ref{fig:other_results_CoT} in Appendix~\ref{sec:Chain-of-Thought Does Not Alleviate Drift-Induced Degradation}). \textit{\textbf{Overall, the observed effects of context drift persist under these alternative settings, supporting the robustness and generalisability of our conclusions.}}

We find from Figure~\ref{fig:olmo_results}, Figure~\ref{fig:other_results}, and Figure~\ref{fig:slope} that \textit{\textbf{the impact of natural context drift is more pronounced in benchmarks that emphasize surface-form alignment, whereas tasks requiring deeper reasoning exhibit greater resilience.}} The downward trend is less pronounced for datasets requiring advanced reasoning, such as \textsc{WikiWhy} and \textsc{HotpotQA}. This may be because natural text evolution introduces diverse linguistic cues and contextual variations that activate broader reasoning mechanisms in LLMs, partially offsetting the impact of semantic drift. In contrast, surface-level QA tasks such as \textsc{SQuAD} rely more heavily on lexical or structural cues \citep{schlegel-etal-2020-framework, wu-etal-2021-understanding} and therefore appear more sensitive to the text evolution. Dataset-level statistics support this contrast: \textsc{BoolQ} and \textsc{SQuAD 2.0} exhibit quite steep average slopes (67.23 and 83.50, respectively) and strong Pearson correlations between average accuracy and semantic similarity (e.g., \textsc{BoolQ}'s average correlation = 0.867 ± 0.064). Meanwhile, \textsc{HotpotQA} demonstrates a much shallower 23.94 average slope and lower correlation, indicating greater robustness to textual edits.

\begin{figure}
    \centering
    \includegraphics[width=\linewidth]{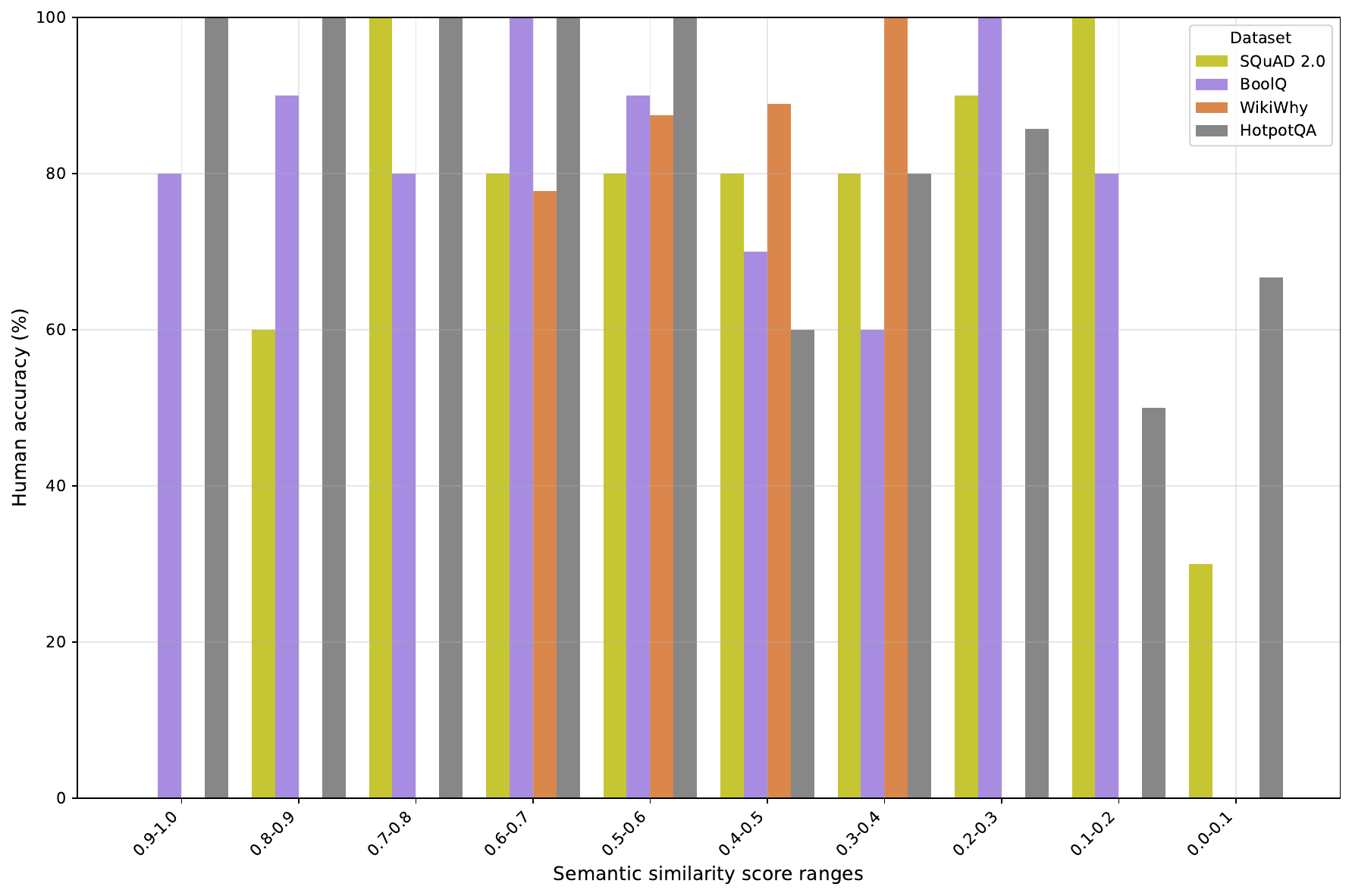}
    \caption{Accuracy of human annotators on QA tasks across semantic similarity bins.}
    \label{fig:humanQAaccuracy}
\end{figure}

\textit{\textbf{Unlike LLMs, human performance in reading comprehension is not influenced by deviations in measured semantic similarity.}} To determine whether the observed decline in LLMs’ accuracy is due to reduced semantic similarity or the possibility that the edited reading paragraphs became degraded and unanswerable, we evaluate human performance across semantic similarity bins within the four investigated datasets. For each dataset, we randomly sample an equal number of edited QA instances from each bin, assign two annotators to label the answers, and involve a third annotator to resolve any disagreements. Details of the human annotation protocol are provided in Appendix~\ref{sec:Human Annotation Details}. As shown in Figure~\ref{fig:humanQAaccuracy}, across all QA benchmarks, human performance does not consistently decline with decreasing semantic similarity, supporting the conclusion that the degradation in LLM accuracy stems from semantic drift rather than a loss of question answerability.

\textit{\textbf{There may be little concern regarding the impact of paragraph leakage from other data sources beyond Wikipedia.}} A natural question arises as to whether, given the vast scale of LLMs' training corpora, the edited versions of the reading paragraphs might also appear in sources beyond the Wikipedia subset we focus on, potentially affecting the findings. Therefore, using the infini-gram engine~\citep{liu2024infinigram}, we aim to estimate as accurately as possible the percentage of edited reading paragraphs that appear verbatim in the complete training corpus of the evaluated LLMs across the six examined QA benchmarks, as shown in Table~\ref{tab:other_sources}. Infini-gram enables efficient querying over the whole training data of \texttt{OLMo-2-1124-13B-Instruct} (whose results are also used as a proxy for \texttt{OLMo-2-1124-7B-Instruct}, given the same training data shared) and \texttt{OLMoE-1B-7B-0125-Instruct}. For the remaining LLMs, we are limited to querying their pretraining corpora (which already account for a significant portion of the overall training data): \textsc{Dolma 1.7} for \texttt{OLMo-7B-0724-Instruct-hf}, \textsc{RedPajama v1} for \texttt{AmberChat} and \texttt{TinyLlama-1.1B-Chat-v1.0} (used as a proxy), \textsc{Pile} for \texttt{dolly-v2-7b} and \texttt{dolly-v2-12b}. As shown in Table~\ref{tab:other_sources}, verbatim inclusion of the edited paragraphs in the LLMs’ training corpora remains negligible across the board. With additional consideration that our analysis does not include all the edited reading paragraphs, we believe that the impact of these paragraphs appearing in other data sources may not be significant. Finally, we emphasise that our methodology focuses on measuring the semantic similarity between the edited reading paragraphs and the content from the same source, i.e., Wikipedia. Therefore, extending the analysis to other potential sources falls outside the scope of this paper, and we leave this for future investigation.

\begin{table}[htb!]
    \centering
    \resizebox{\linewidth}{!}{
    \begin{tabular}{lcccccc}
    \toprule
    &\textsc{SQuAD 1.1}&\textsc{SQuAD 2.0}&\textsc{D(RoBERTa)}&\textsc{BoolQ}&\textsc{WikiWhy}&\textsc{HotpotQA}\\
    \midrule
    \texttt{OLMo-7B-0724-Instruct-hf}&1.59&3.94&1.33&4.30&1.52&-\\
    \texttt{OLMo-2-1124-7B/13B-Instruct}&4.85&8.53&4.79&2.75&1.14&-\\
    \texttt{OLMoE-1B-7B-0125-Instruct}&4.77&8.69&4.79&2.75&1.14&-\\
    \texttt{AmberChat}\&\texttt{TinyLlama-1.1B-Chat-v1.0}&1.04&2.40&2.13&4.47&1.14&-\\
    \texttt{dolly-v2-7b/12b}&0.18&1.07&-&0.40&0.76&-\\
    \bottomrule
    \end{tabular}}
    \caption{Percentage (\%) of edited reading paragraphs that appear verbatim in the training corpora of the evaluated LLMs.}
    \label{tab:other_sources}
\end{table}

\section{Conclusion}
\label{sec:Conclusion}

We introduce a novel methodology for examining how real-world natural context evolution affects the language understanding of LLMs when it deviates semantically from their training data. Leveraging Wikipedia revision histories, we curate naturally human-edited variants of benchmark reading passages, compute their semantic similarity to versions present in the models' training corpus, and correlate these similarity scores with model performance. Our empirical findings show that, while natural text evolution has little to no effect on human QA performance, LLMs generally exhibit a consistent downward trend in accuracy as semantic similarity decreases. We hope this study contributes to the growing body of research focused on understanding and addressing the limitations of LLMs in real-world, evolving textual contexts.

\section*{Limitations}

Our work has several limitations. First, we focus exclusively on the QA task; extending these findings to other downstream tasks remains an open avenue for future research. Second, we evaluate only transparent LLMs with fully open-access training corpora. While this constraint is necessary for our methodology, some of these models still lag behind state-of-the-art proprietary LLMs in performance. Applying our framework to such proprietary models would be a valuable extension, although it would require access to their training data, which is currently unavailable.

\bibliography{latex/anthology,latex/custom}

\appendix

\section{Answer Preservation Check and Data Statistics}
\label{sec:Answer Preservation Check and Data Statistics}

For answer preservation checking, in the extractive QA setting (including the extractive question set for \textsc{HotpotQA}), we ensure that at least one (or all) of the ground truth answers can still be found in the edited passage. For \textsc{BoolQ}, we manually inspect the generated edited test set and remove instances where the edited passage contains fewer than 56 characters. We also check \textsc{WikiWhy}, but no filtering is applied. Table~\ref{tab:statistics} presents the statistics of the extracted data for each QA dataset.

\begin{table}[htb!]
\centering
\resizebox{\columnwidth}{!}{
\begin{tabular}{lccccc}
\toprule
\textbf{Dataset} & \textbf{Titles} & \textbf{Passages} & \textbf{Edited} & \textbf{Avg. per} & \textbf{Questions} \\
 & \textbf{(ext./total)} & & \textbf{Passages} & \textbf{Passage} & \\
\midrule
\textsc{SQuAD 1.1} \citep{rajpurkar-etal-2016-squad} & 47/48 & 825 & 1,531 & 8.85 & 3,920 \\
\midrule
\textsc{SQuAD 2.0} \citep{rajpurkar-etal-2018-know} & 47/48 & 466 & 914 & 17.82 & 4,281 \\
\midrule
\textsc{D(RoBERTa)} \citep{bartolo-etal-2020-beat} & 47/48 & 135 & 249 & 5.56 & 376 \\
\midrule
\textsc{BoolQ} \citep{clark-etal-2019-boolq} & 2,488/2,651 & 957 & 2,559 & 3.16 & 1,064 \\
\midrule
\textsc{WikiWhy} \citep{ho2023wikiwhy} & 833/873 & 193 & 251 & 1.36 & 193 \\
\midrule
\textsc{HotpotQA} \citep{yang-etal-2018-hotpotqa} & 10,971/13,783 & 2,948 & 7,350 & 2.79 & 2,970 \\
\bottomrule
\end{tabular}}
\caption{Summary of QA benchmarks with paragraph evolution histories extracted from Wikipedia edits.}
\label{tab:statistics}
\end{table}

\section{Percentage of Instances where LLMs Succeed on Context-Free Question Answering}
\label{sec:Percentage of Instances where LLMs Succeed on Context-Free Question Answering}

\begin{table}[htb!]
\centering
\resizebox{\columnwidth}{!}{
\begin{tabular}{lcccc}
\toprule
\diagbox{\textbf{Dataset}}{\textbf{LLM}} & \texttt{OLMo-7B-0724-Instruct-hf} & \texttt{OLMo-2-1124-7B-Instruct} & \texttt{OLMo-2-1124-13B-Instruct} & \texttt{OLMoE-1B-7B-0125-Instruct} \\
\midrule
\textsc{SQuAD 1.1} & $2.63$ & $5.41$ & $5.81$ & $2.42$ \\
\textsc{SQuAD 2.0} & $53.55$ & $52.07$ & $56.48$ & $55.43$ \\
\textsc{D(RoBERTa)} & $2.13$ & $12.52$ & $8.66$ & $6.26$ \\
\textsc{BoolQ} & $36.35$ & $48.69$ & $61.40$ & $45.85$ \\
\textsc{WikiWhy} & $0.00$ & $3.42$ & $0.00$ & $2.66$ \\
\textsc{HotpotQA} & $2.77$ & $9.30$ & $7.89$ & $8.21$ \\
\bottomrule
\end{tabular}}
\caption{Percentage of instances with questions that are correctly answered by an LLM without access to the context paragraph.}
\label{tab:memorisation_percentage}
\end{table}

\section{Inference Prompts for QA Tasks}
\label{sec:qa_prompts}
This appendix provides the complete prompt templates used for zero-shot inference across all QA datasets evaluated in this study.
\begin{center}
\textbf{\textsc{SQuAD 1.1}} \& \textbf{\textsc{SQuAD 2.0}} \& \textbf{\textsc{D(RoBERTa)}}
\end{center}
\begin{quote}
\textit{Use the provided article delimited by triple quotes to answer question. Provide only the shortest continuous span from the context without any additional explanation. If the question is unanswerable, return ``unanswerable''.}
\texttt{"""{passage}"""}
\textit{Question:} \texttt{question}
\end{quote}
\textit{Rationale:} Explicitly requests the shortest continuous span to encourage precise extraction for extractive reading comprehension tasks.
\begin{quote}
    \textit{\textbf{parametric knowledge testing:} Provide an answer to the given question. If the question is unanswerable, return ``unanswerable''. Do not provide any explanation.}
        \textit{Question:} \texttt{question}
\end{quote}
\begin{center}
\textbf{\textsc{BoolQ}}
\end{center}
\begin{quote}
\textit{Use the provided article delimited by triple quotes to answer question. Return only TRUE or FALSE. If the question is unanswerable, return ``unanswerable''. Do not provide any explanation.}
\texttt{"""{passage}"""}
    \textit{Question:} \texttt{question}
\end{quote}
\textit{Rationale:} Constrains output format to TRUE/FALSE responses for binary classification tasks.
\begin{quote}
    \textit{\textbf{parametric knowledge testing:} Provide an answer to the given question. Return only TRUE or FALSE. If the question is unanswerable, return ``unanswerable''. Do not provide any explanation.}
        \textit{Question:} \texttt{question}
\end{quote}
\begin{center}
\textbf{\textsc{WikiWhy}} \& \textbf{\textsc{HotpotQA}}
\end{center}
\begin{quote}
\textit{Use the provided article delimited by triple quotes to answer question. If the question is unanswerable, return ``unanswerable''. Do not provide any explanation.}
\texttt{"""{passage}"""}
    \textit{Question:} \texttt{question}
\end{quote}
\textit{Rationale:} Allows for more flexible answer generation while maintaining the unanswerable option for cause-effect reasoning and multi-hop reasoning tasks.
\begin{quote}
    \textit{\textbf{parametric knowledge testing:} Provide an answer to the given question. If the question is unanswerable, return ``unanswerable''. Do not provide any explanation.}
        \textit{Question:} \texttt{question}
\end{quote}

\section{Slope Analysis of Accuracy vs. Semantic Similarity}
\label{sec:Slope Analysis of Accuracy vs. Semantic Similarity}

\begin{figure*}[htbp]
   \centering
   
   \begin{subfigure}[b]{0.45\textwidth}
       \centering
       \includegraphics[width=\textwidth]{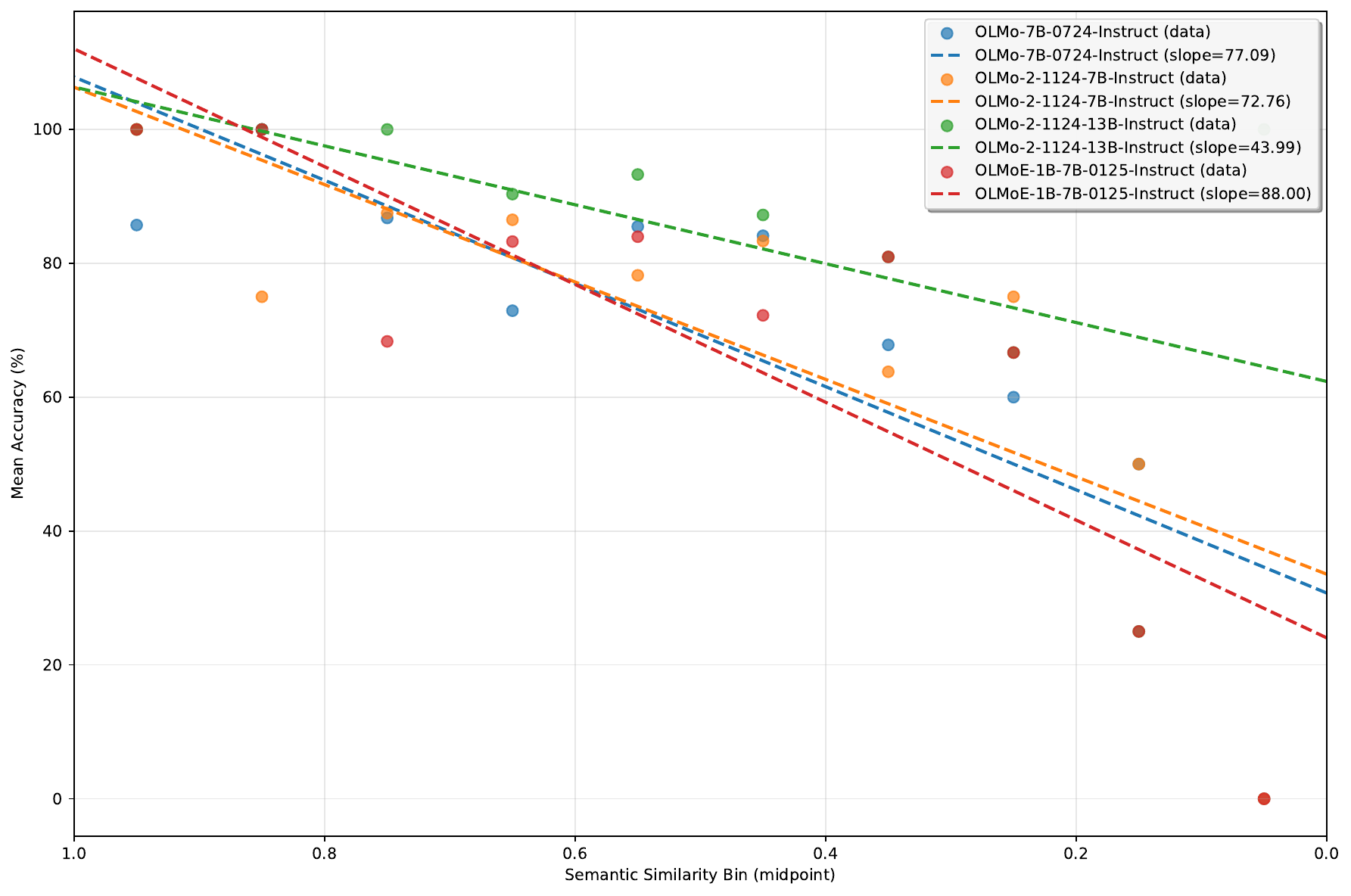}
       \caption{\textsc{SQuAD 1.1}}
   \end{subfigure}
   \hfill
   \begin{subfigure}[b]{0.45\textwidth}
       \centering
       \includegraphics[width=\textwidth]{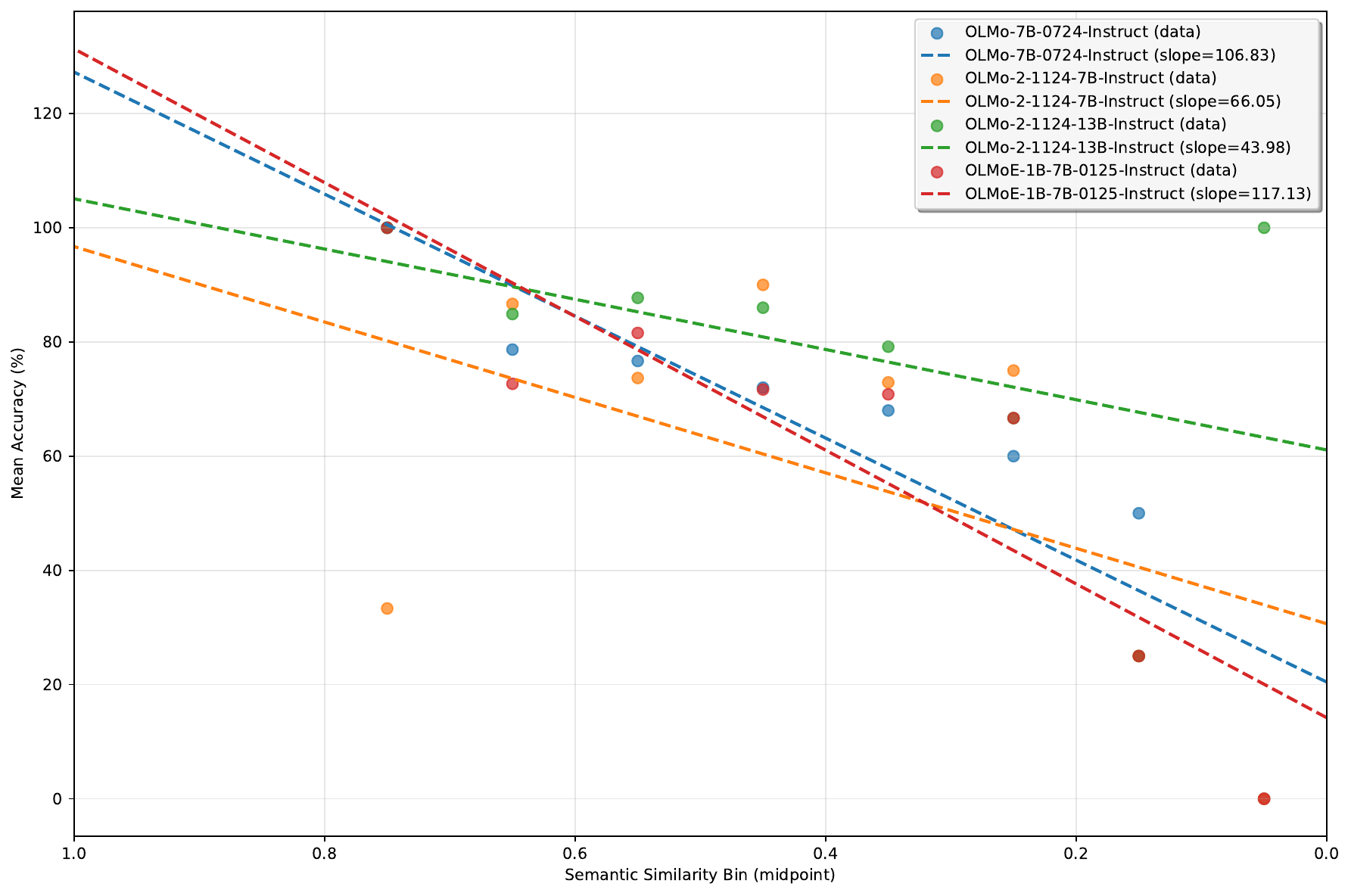}
       \caption{\textsc{SQuAD 2.0}}
   \end{subfigure}
   
   \vspace{1em}
   
   \begin{subfigure}[b]{0.45\textwidth}
       \centering
       \includegraphics[width=\textwidth]{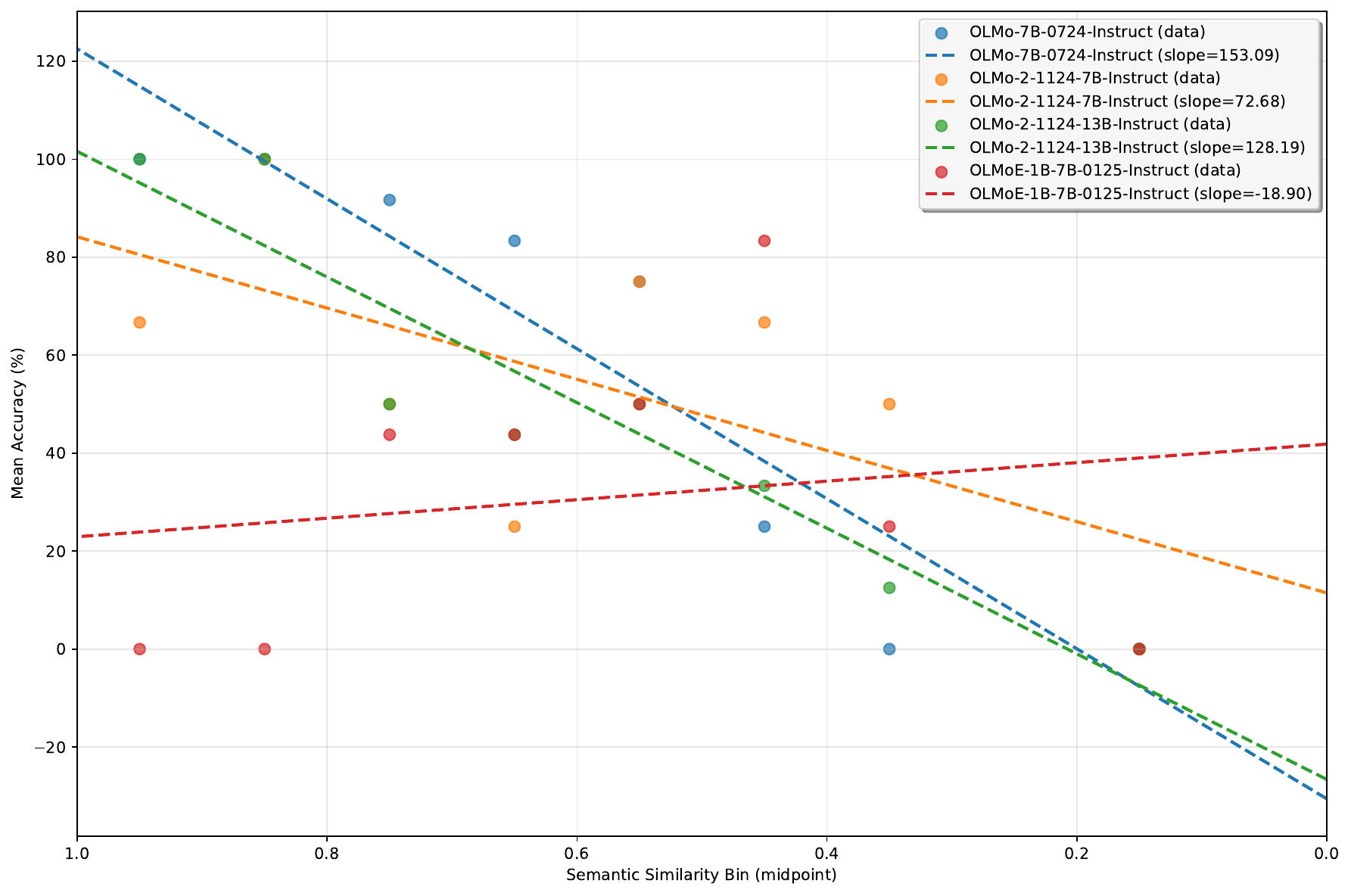}
       \caption{\textsc{D(RoBERTa)}}
   \end{subfigure}
   \hfill
   \begin{subfigure}[b]{0.45\textwidth}
       \centering
       \includegraphics[width=\textwidth]{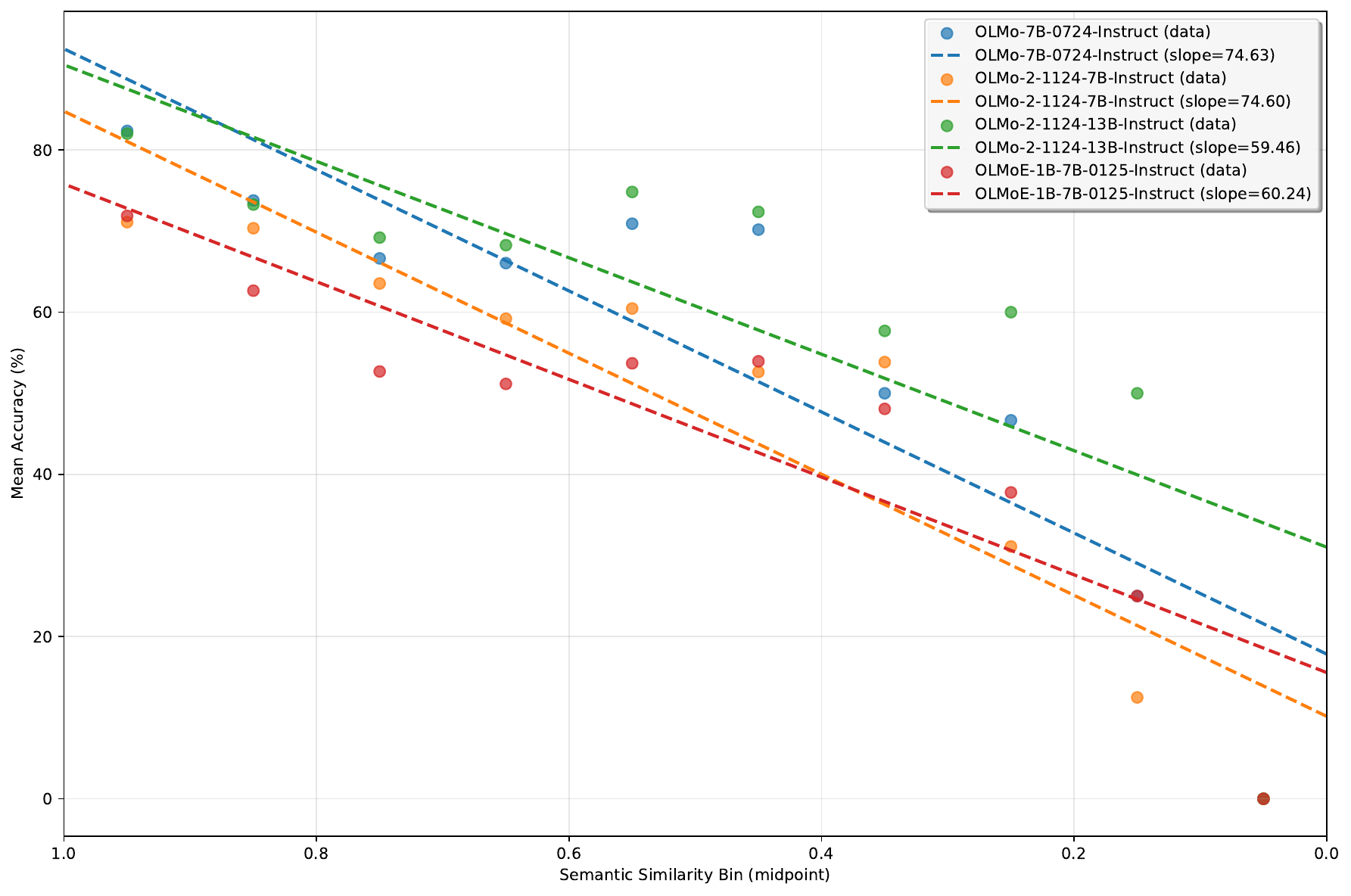}
       \caption{\textsc{BoolQ}}
   \end{subfigure}
   
   \vspace{1em}
   
   \begin{subfigure}[b]{0.45\textwidth}
       \centering
       \includegraphics[width=\textwidth]{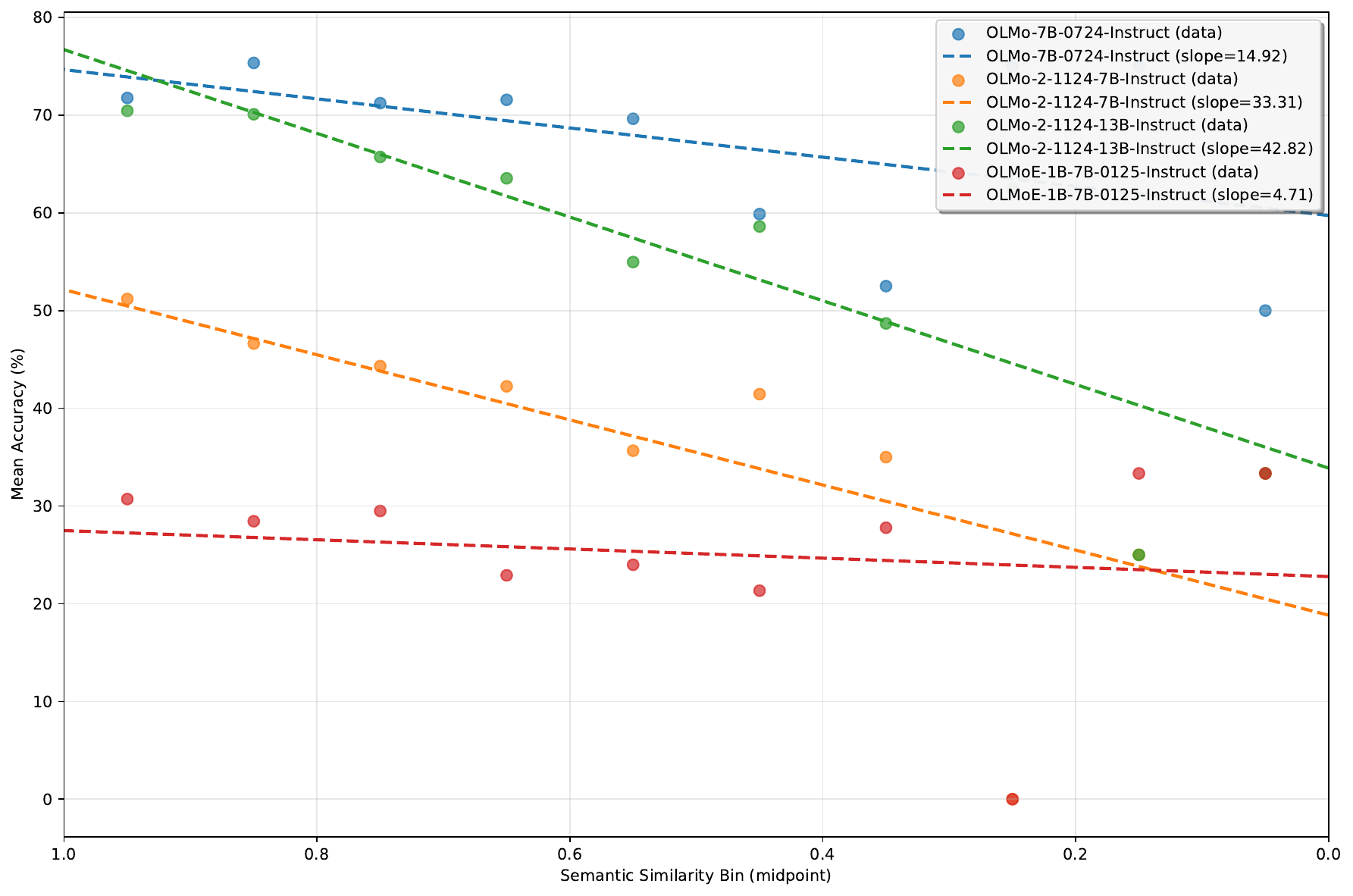}
       \caption{\textsc{HotpotQA}}
   \end{subfigure}
   
   \caption{Average accuracy of instruction-finetuned \texttt{OLMo} LLMs across semantic similarity bins. Each point represents the mean accuracy within a similarity bin, and the dashed lines are linear regression fits summarizing the accuracy trend for each model. Slope values indicate how rapidly model accuracy changes with semantic divergence.}
   \label{fig:slope}
\end{figure*}

\section{Consequences of Context Drift Persist Across Other Embedding Models}
\label{sec:Consequences of Context Drift Persist Across Other Embedding Models}

\begin{figure*}[htb!]
\centering

\begin{subfigure}[b]{0.3\textwidth}
\includegraphics[width=\textwidth]{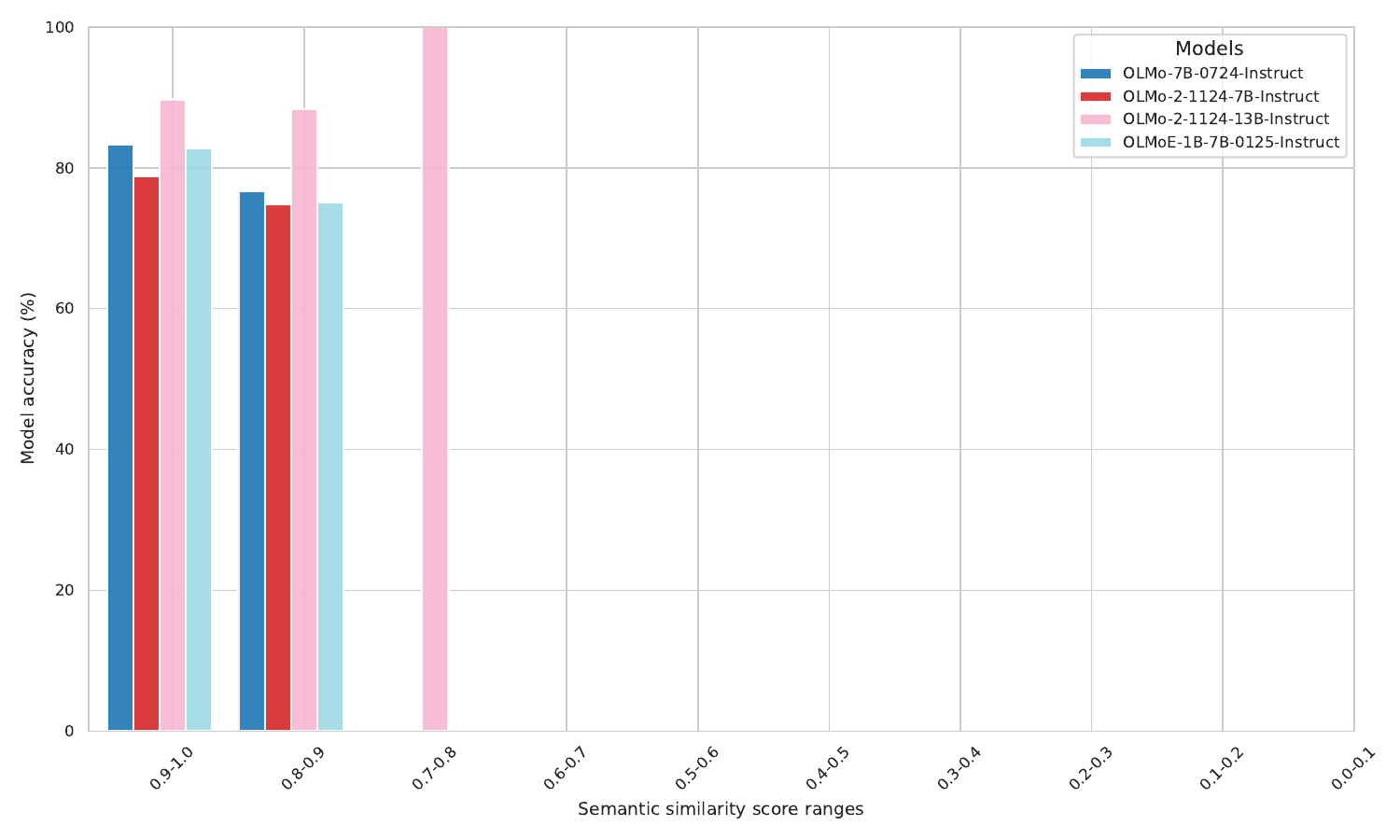}
\caption{\textsc{SQuAD 1.1}}
\end{subfigure}
\hfill
\begin{subfigure}[b]{0.3\textwidth}
\includegraphics[width=\textwidth]{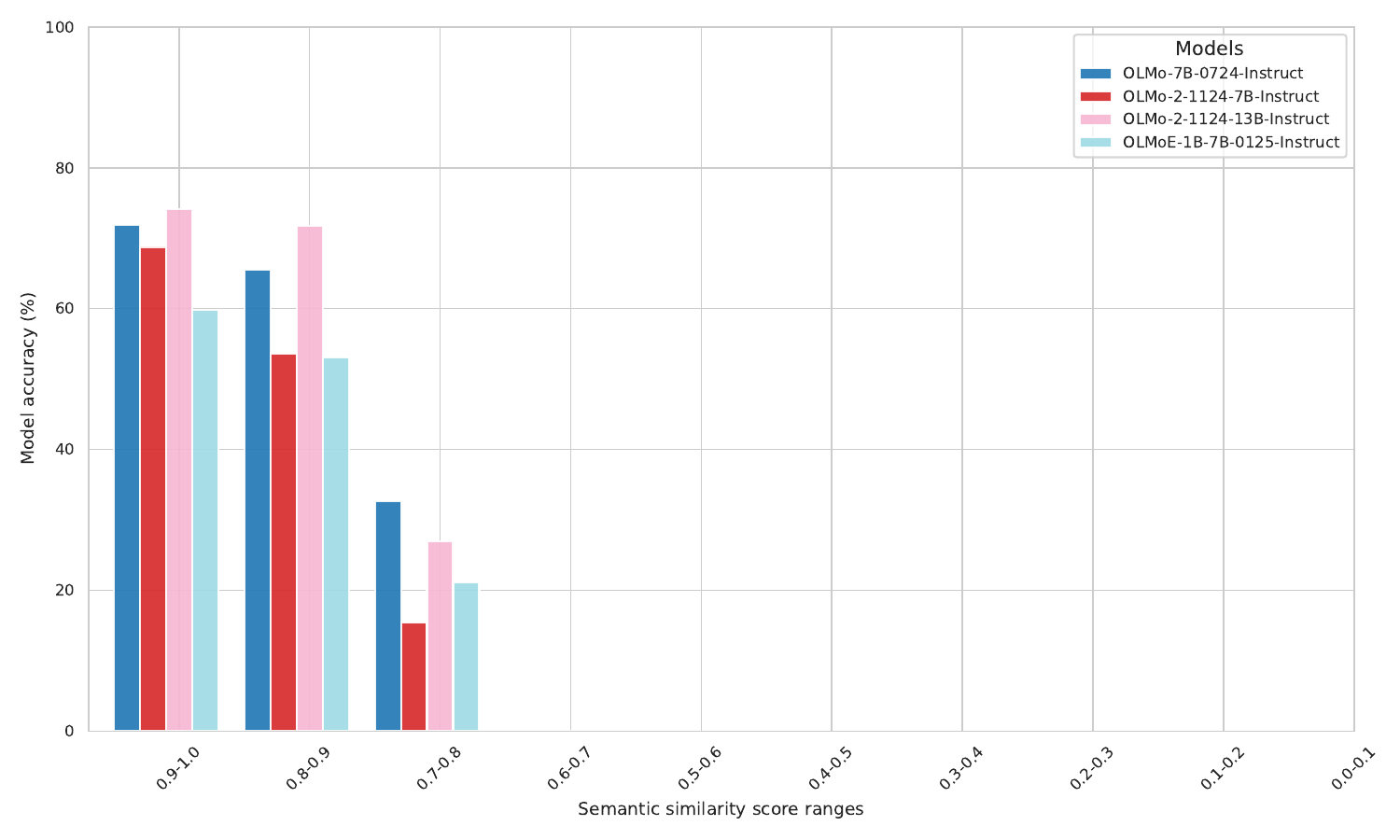}
\caption{\textsc{BoolQ}}
\end{subfigure}
\hfill
\begin{subfigure}[b]{0.3\textwidth}
\includegraphics[width=\textwidth]{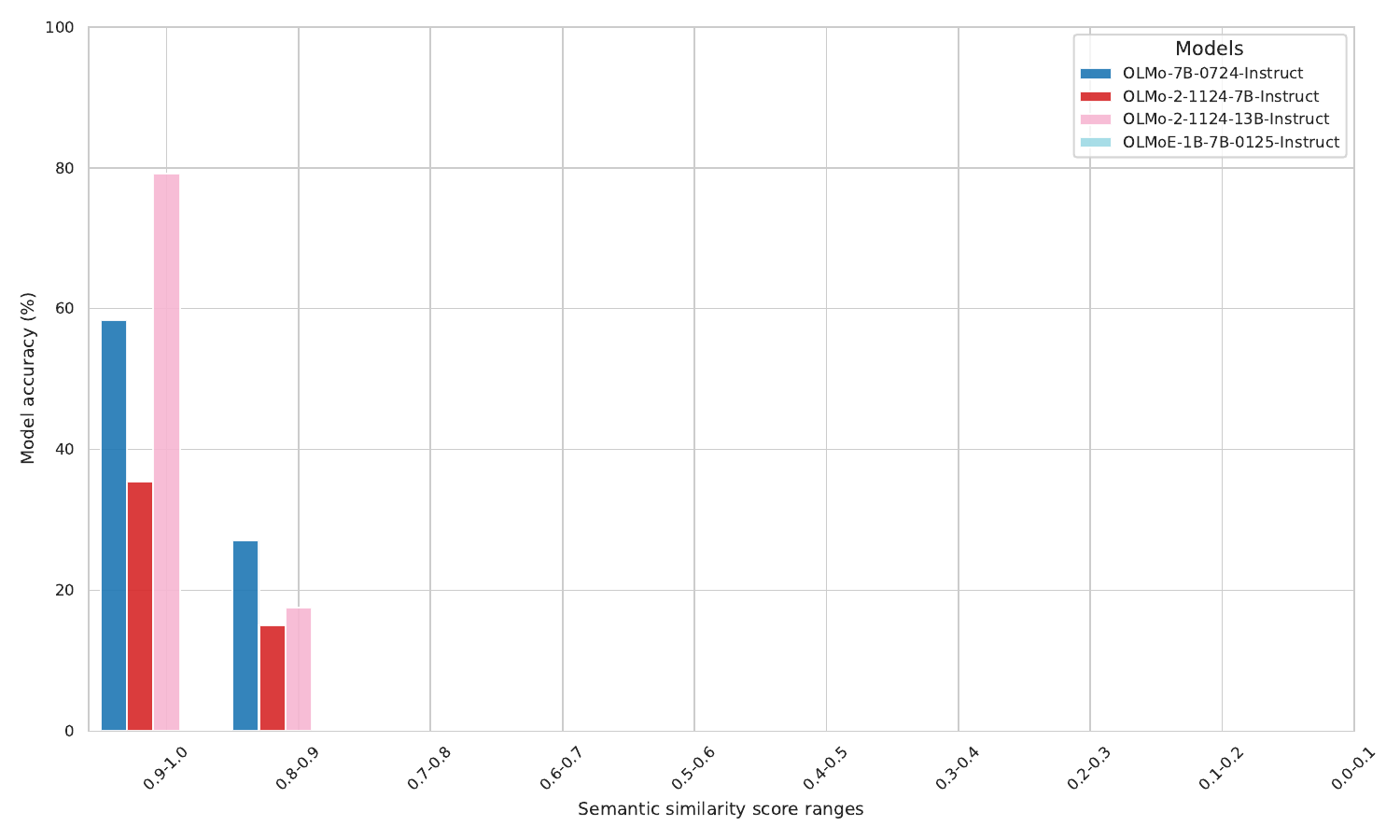}
\caption{\textsc{WikiWhy}}
\end{subfigure}

\vspace{0.5cm}

\begin{subfigure}[b]{0.3\textwidth}
\includegraphics[width=\textwidth]{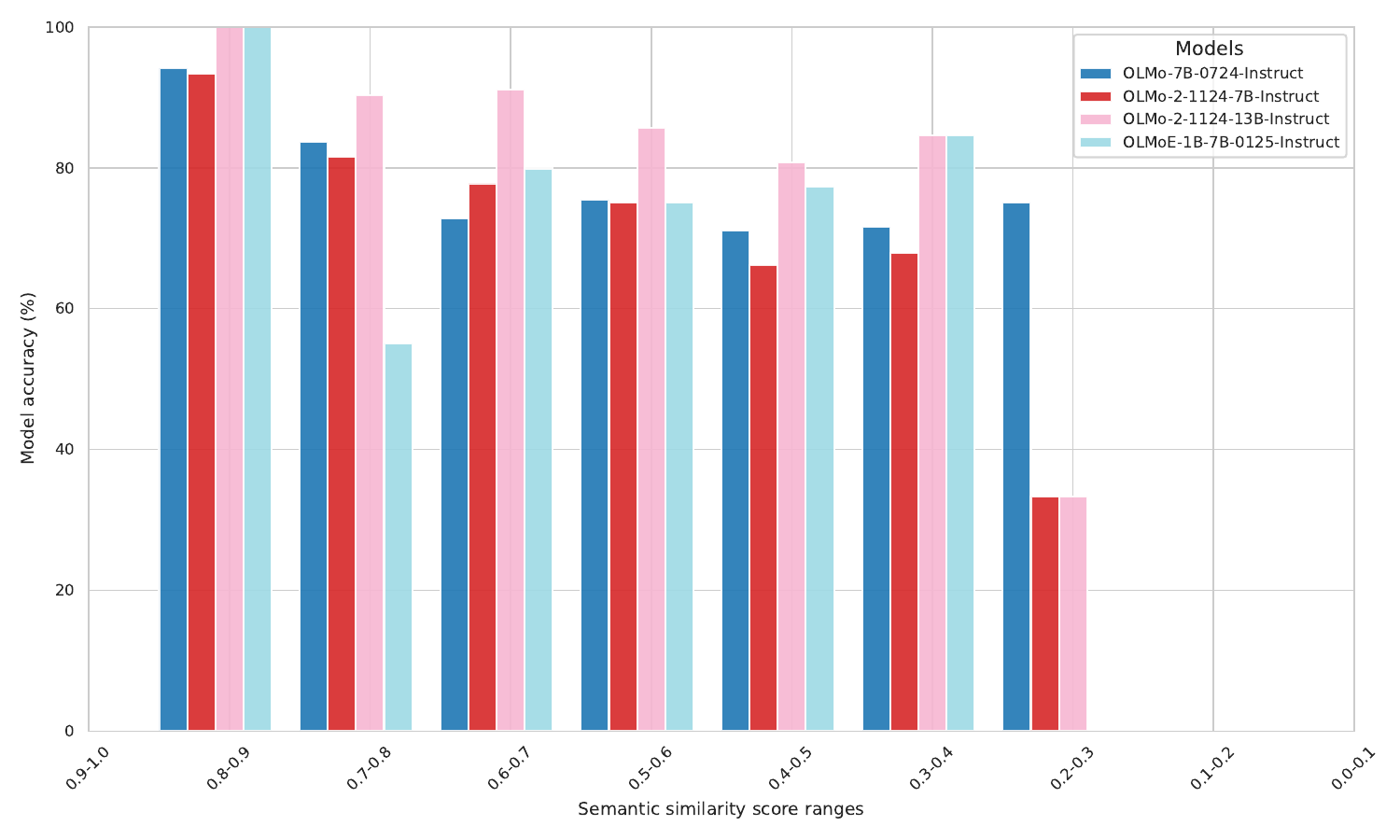}
\caption{\textsc{SQuAD 1.1}}
\end{subfigure}
\hfill
\begin{subfigure}[b]{0.3\textwidth}
\includegraphics[width=\textwidth]{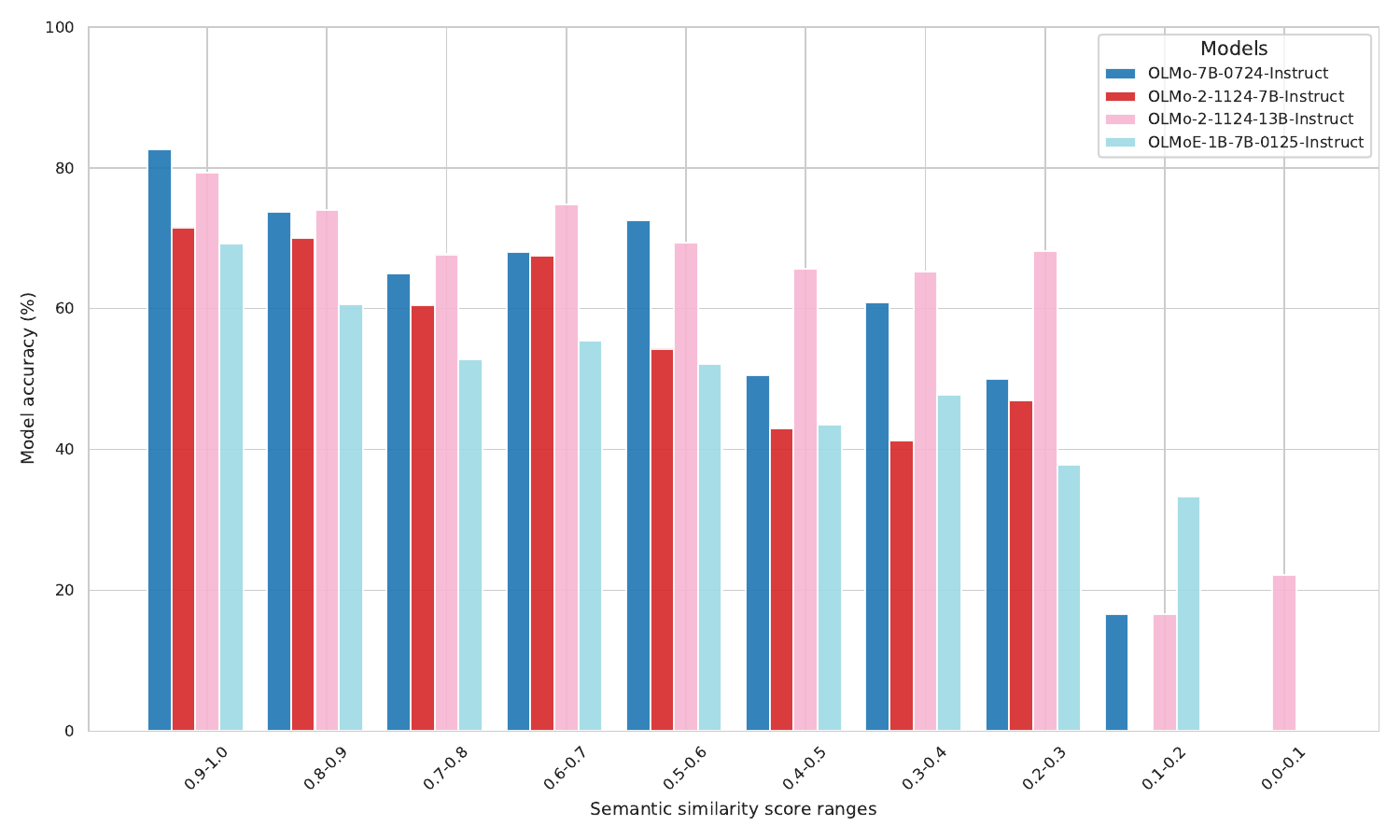}
\caption{\textsc{BoolQ}}
\end{subfigure}
\hfill
\begin{subfigure}[b]{0.3\textwidth}
\includegraphics[width=\textwidth]{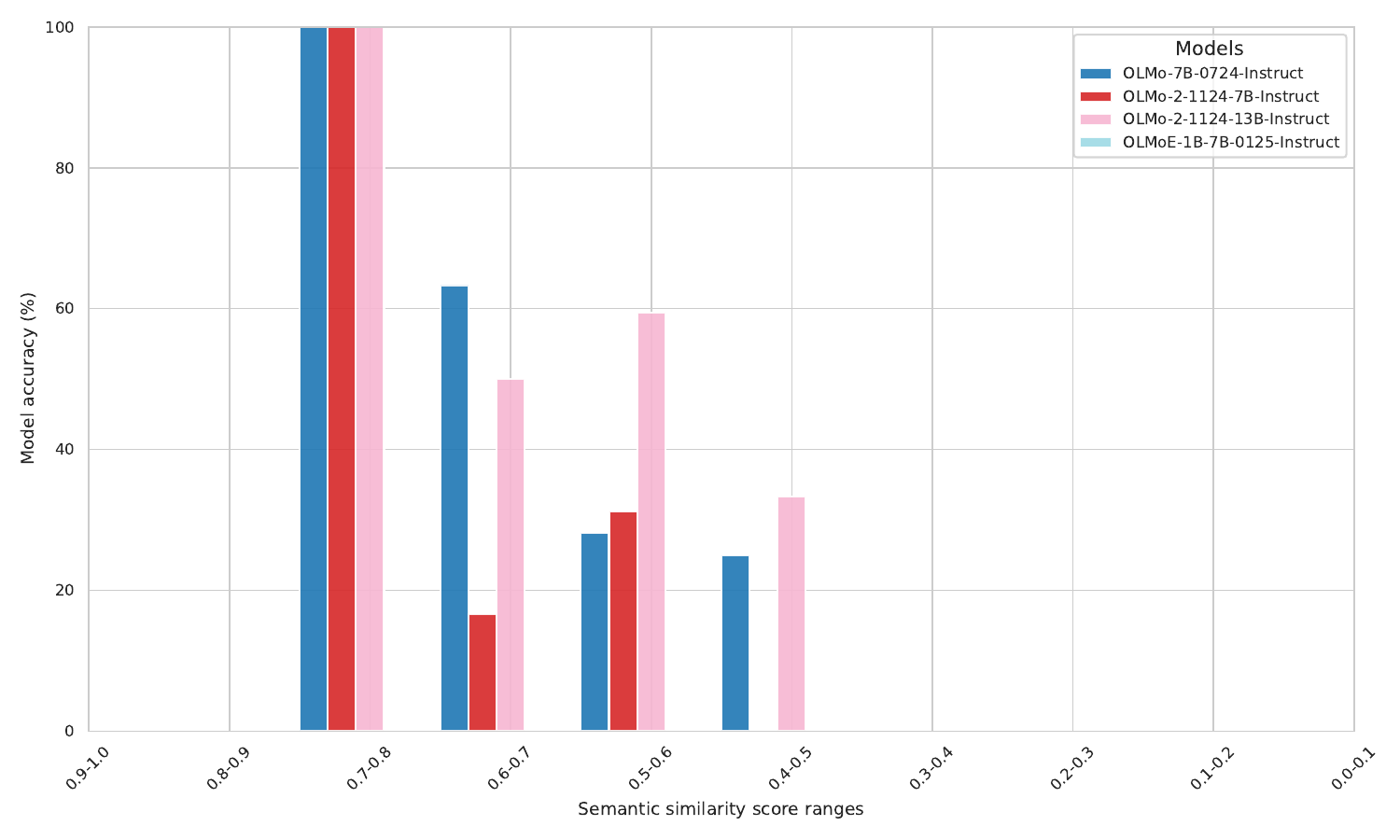}
\caption{\textsc{WikiWhy}}
\end{subfigure}

\vspace{0.5cm}

\begin{subfigure}[b]{0.3\textwidth}
\includegraphics[width=\textwidth]{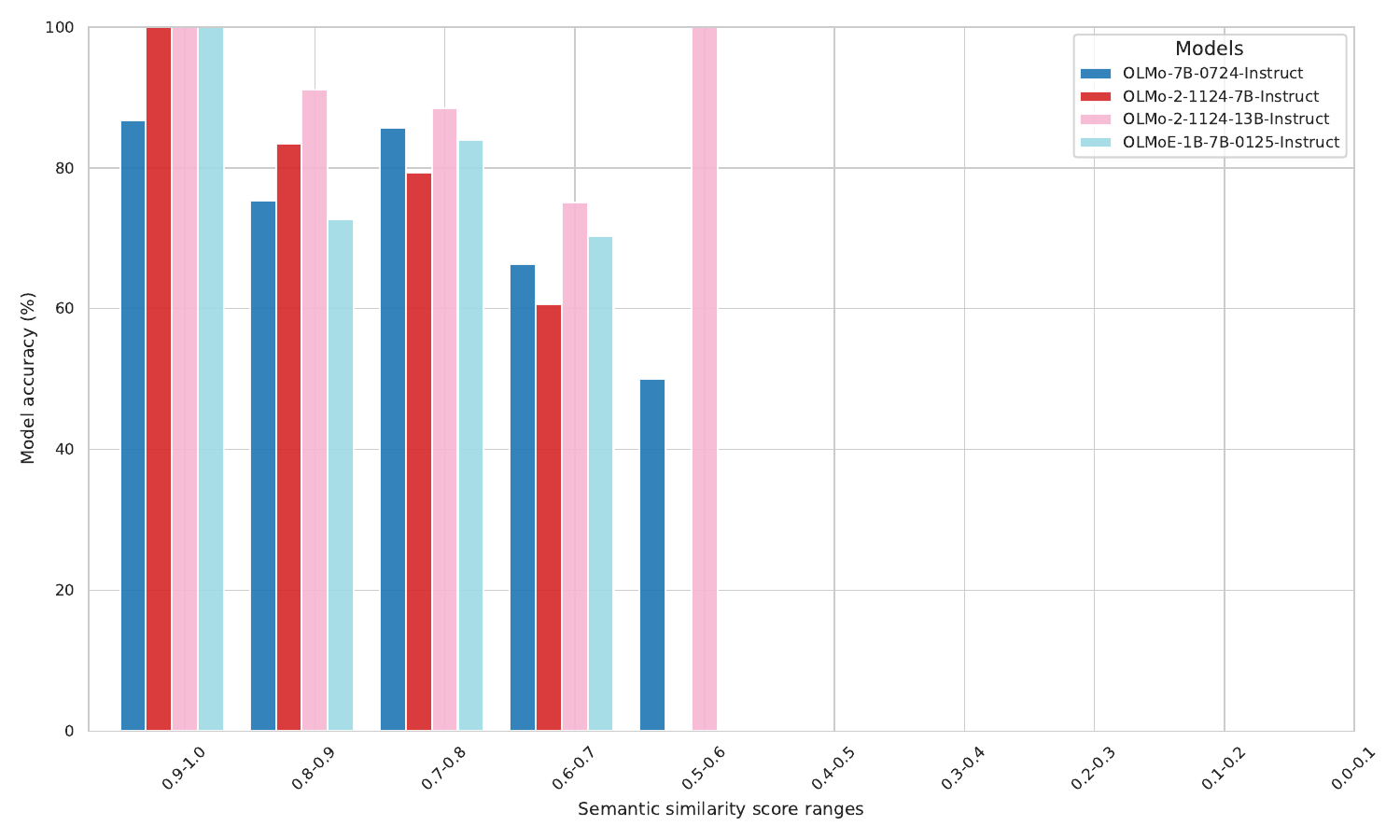}
\caption{\textsc{SQuAD 1.1}}
\end{subfigure}
\hfill
\begin{subfigure}[b]{0.3\textwidth}
\includegraphics[width=\textwidth]{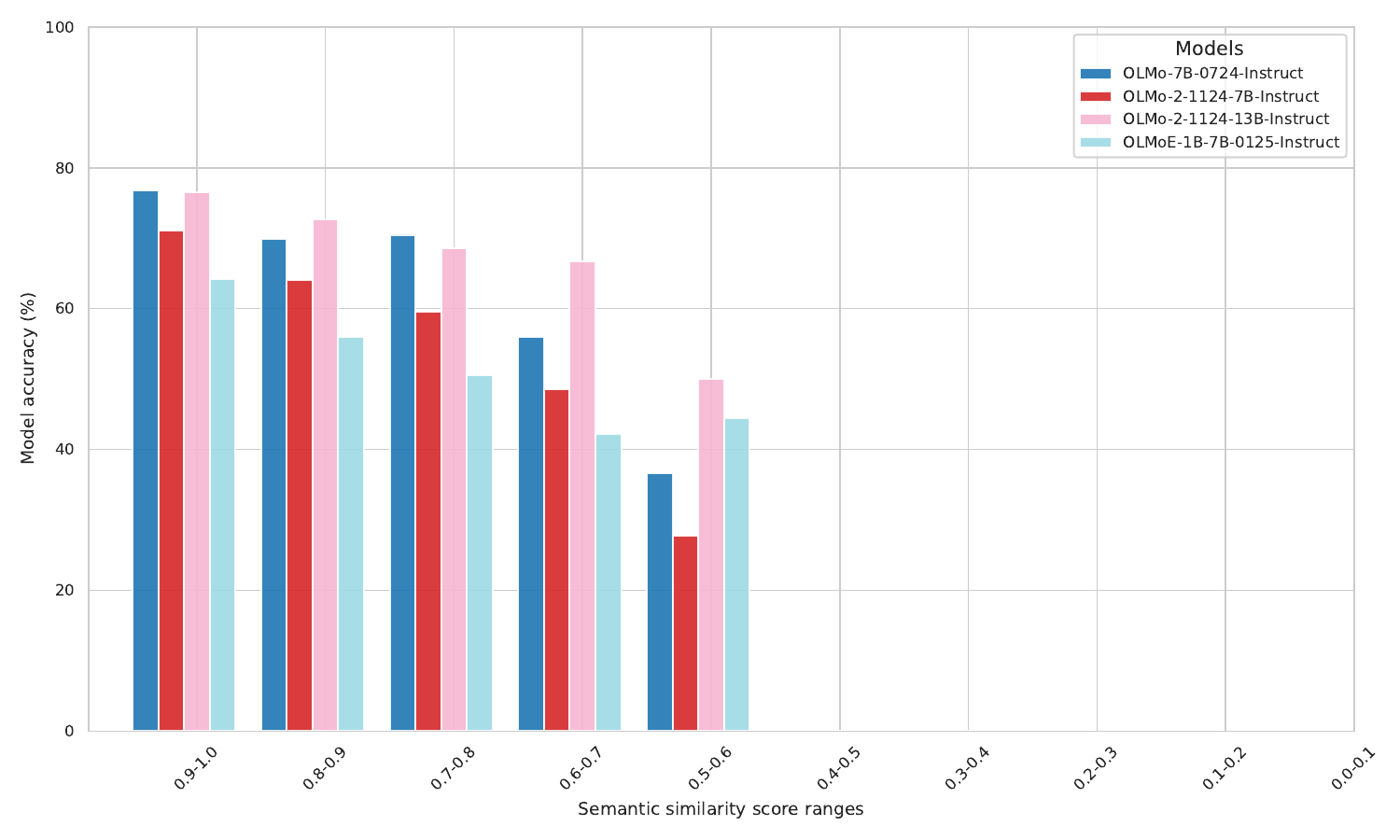}
\caption{\textsc{BoolQ}}
\end{subfigure}
\hfill
\begin{subfigure}[b]{0.3\textwidth}
\includegraphics[width=\textwidth]{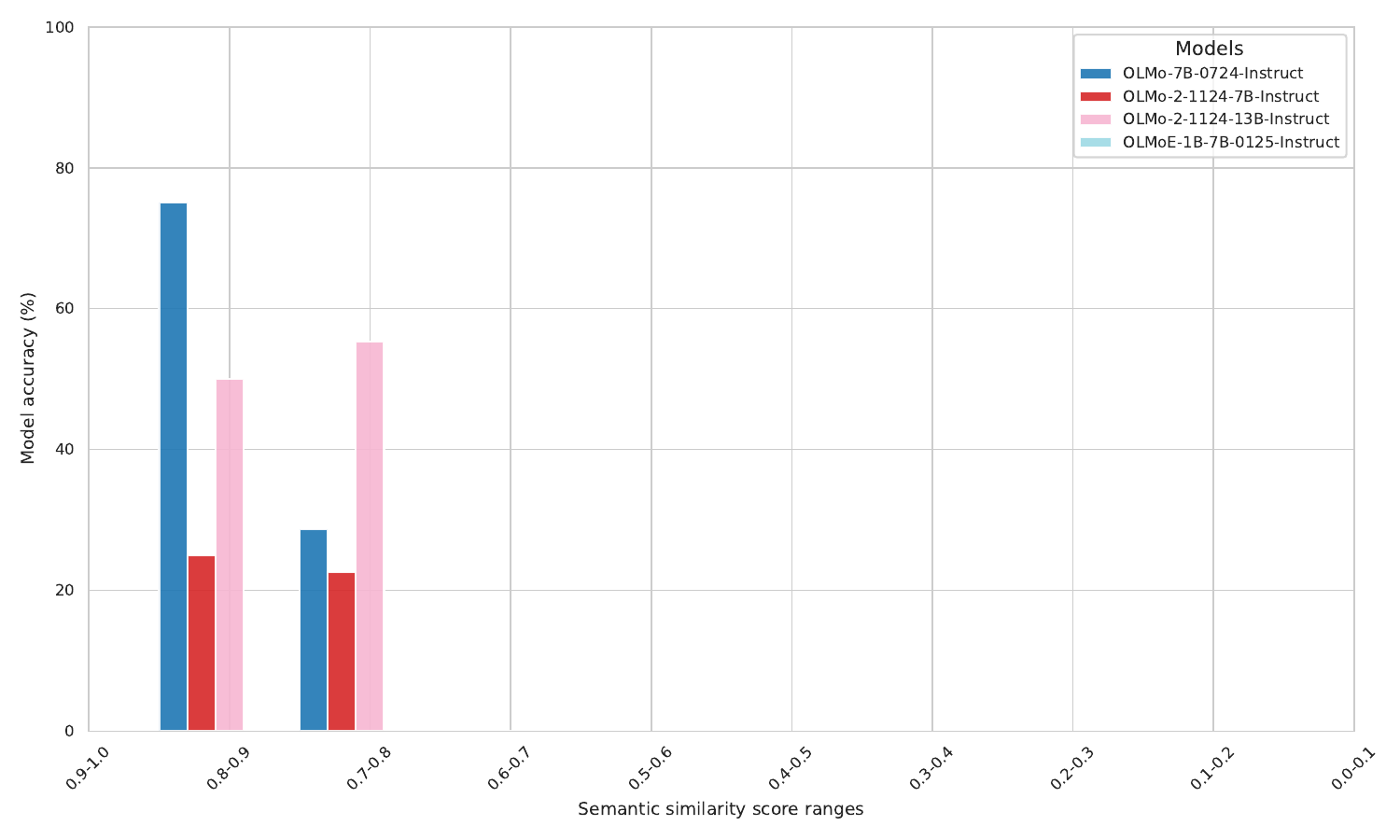}
\caption{\textsc{WikiWhy}}
\end{subfigure}

\caption{Average accuracy of the instruction-finetuned \texttt{OLMo} LLMs across ten semantic similarity bins. The first, second, and third rows show the results obtained using the embedding models \texttt{sentence-t5-base}, \texttt{all-mpnet-base-v2}, and \texttt{bge-small-en-v1.5}, respectively.}
\label{fig:olmo_results_alternative_metrics}
\end{figure*}

\section{Context Drift Effects Hold Without Instruction Tuning}
\label{sec:Context Drift Effects Hold Without Instruction Tuning}

\begin{figure*}[htb!]
    \centering
    \begin{subfigure}[b]{0.3\textwidth}
        \includegraphics[width=\textwidth]{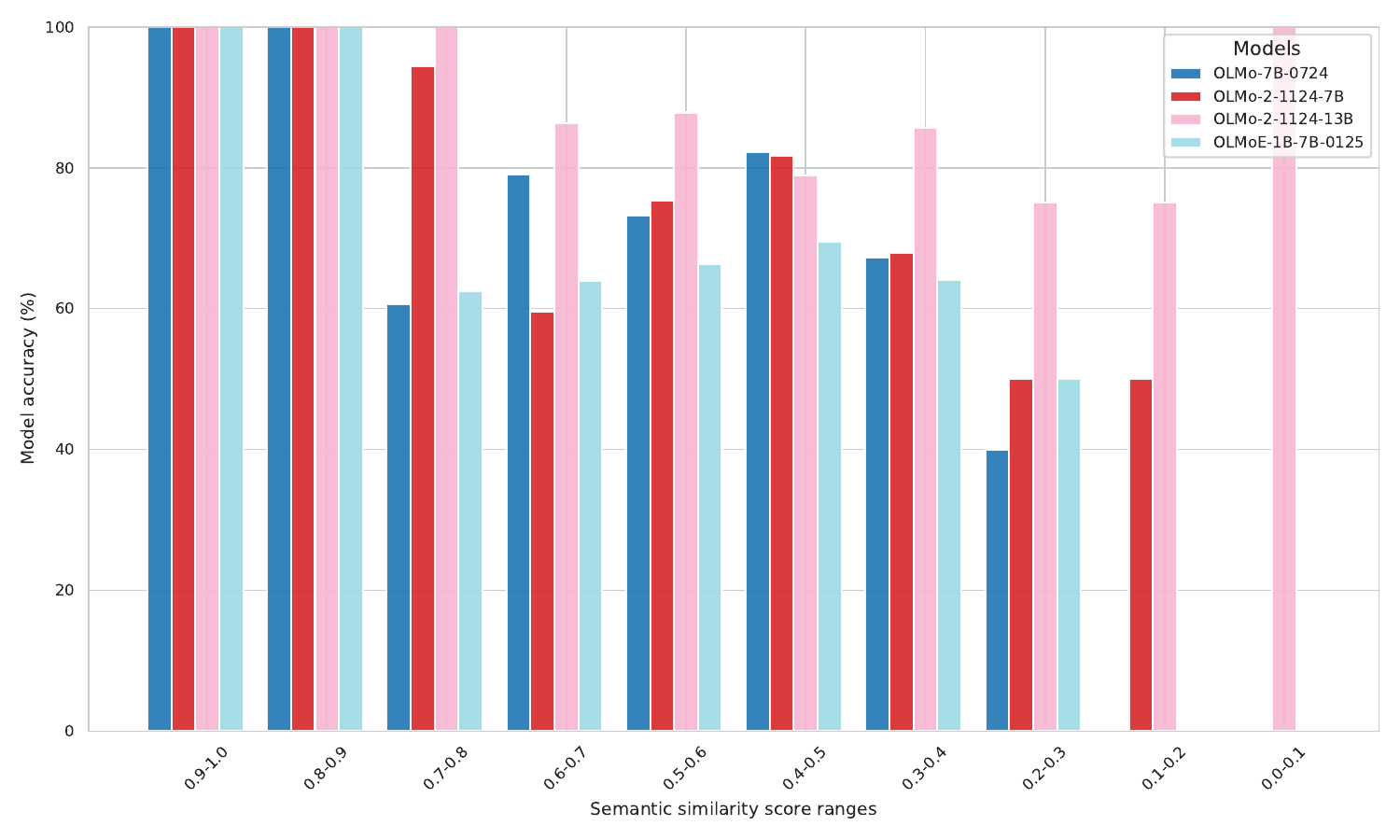}
        \caption{\textsc{SQuAD 1.1}}
    \end{subfigure}
    \hfill
    \begin{subfigure}[b]{0.3\textwidth}
        \includegraphics[width=\textwidth]{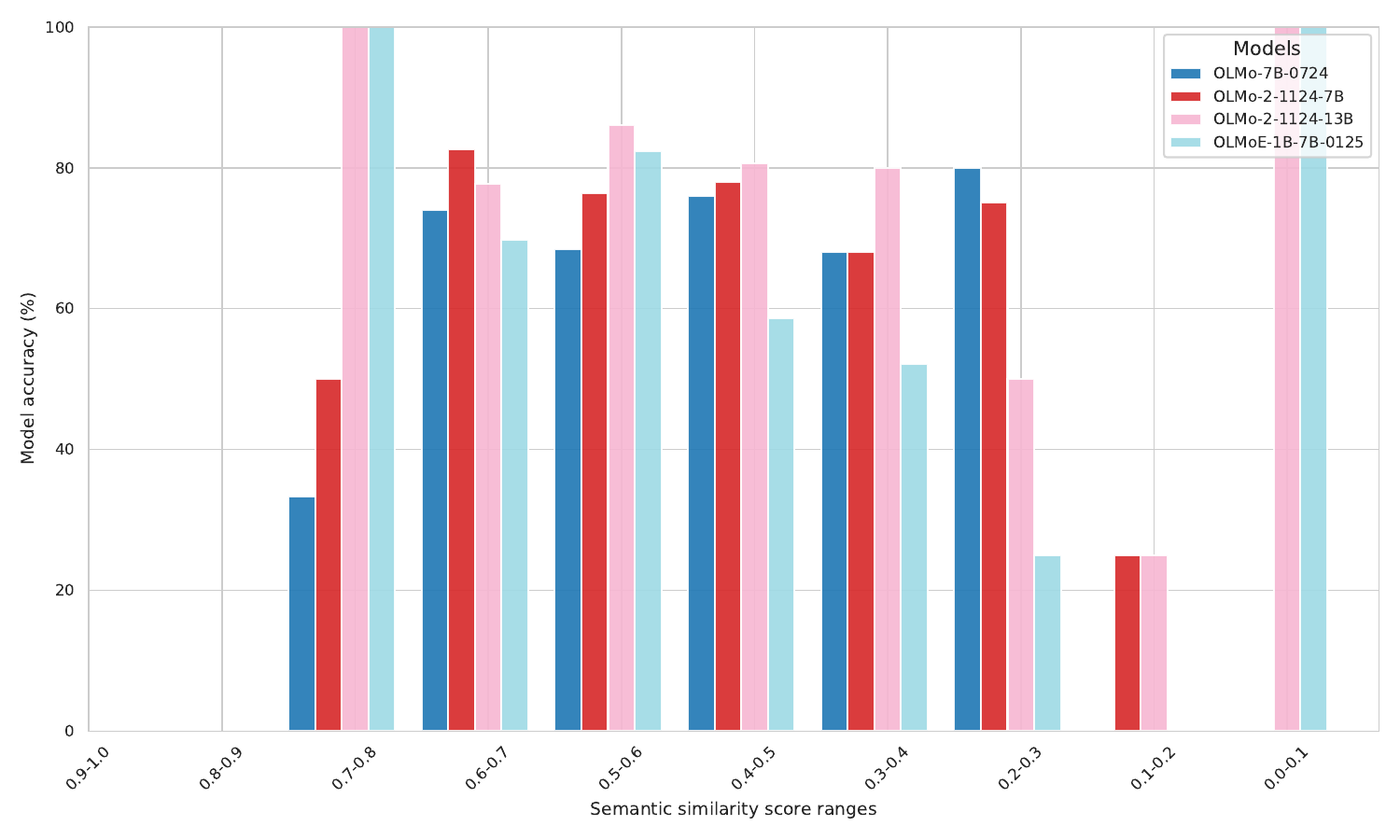}
        \caption{\textsc{SQuAD 2.0}}
    \end{subfigure}
    \hfill
    \begin{subfigure}[b]{0.3\textwidth}
        \includegraphics[width=\textwidth]{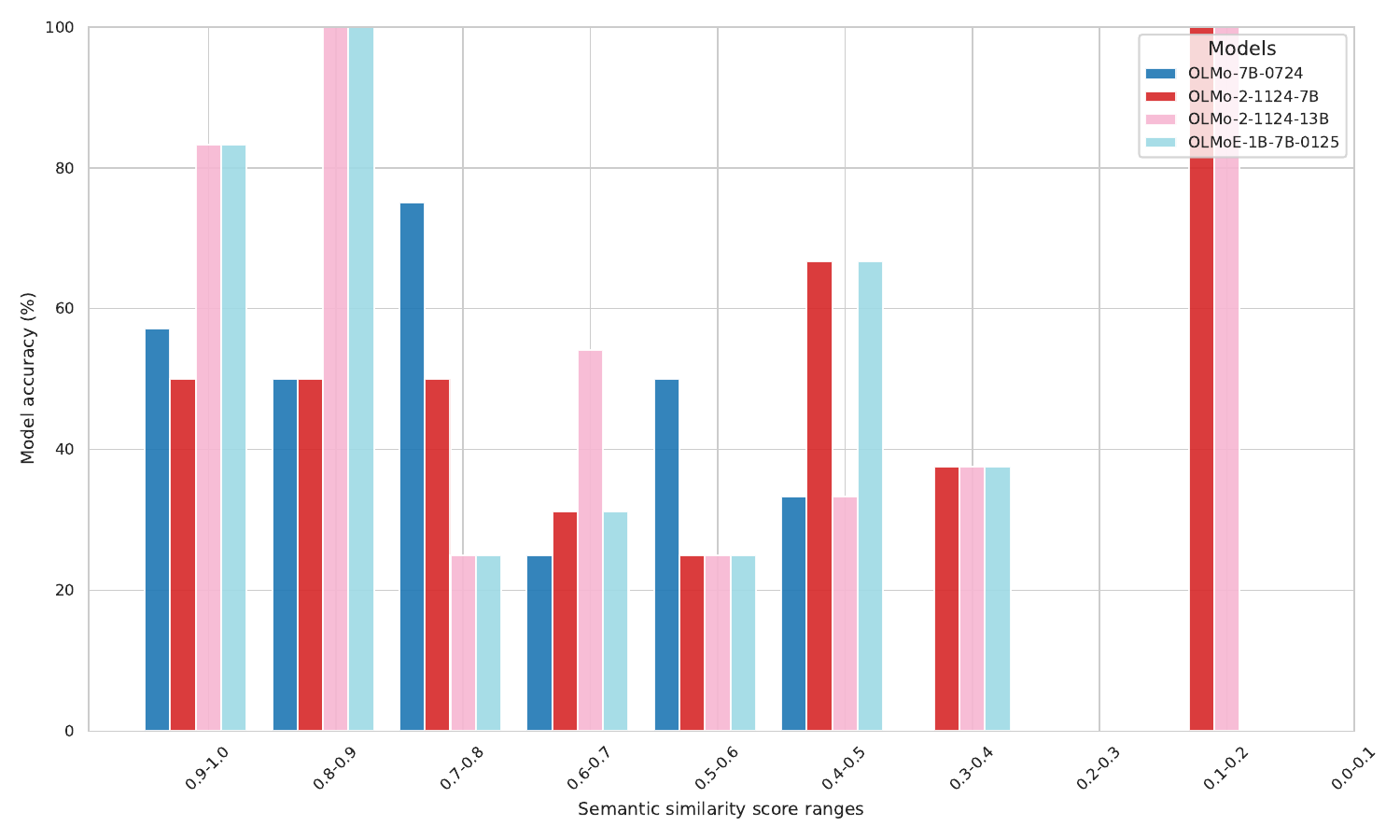}
        \caption{\textsc{D(RoBERTa)}}
    \end{subfigure}

    \vspace{0.5cm}

    \begin{subfigure}[b]{0.3\textwidth}
        \includegraphics[width=\textwidth]{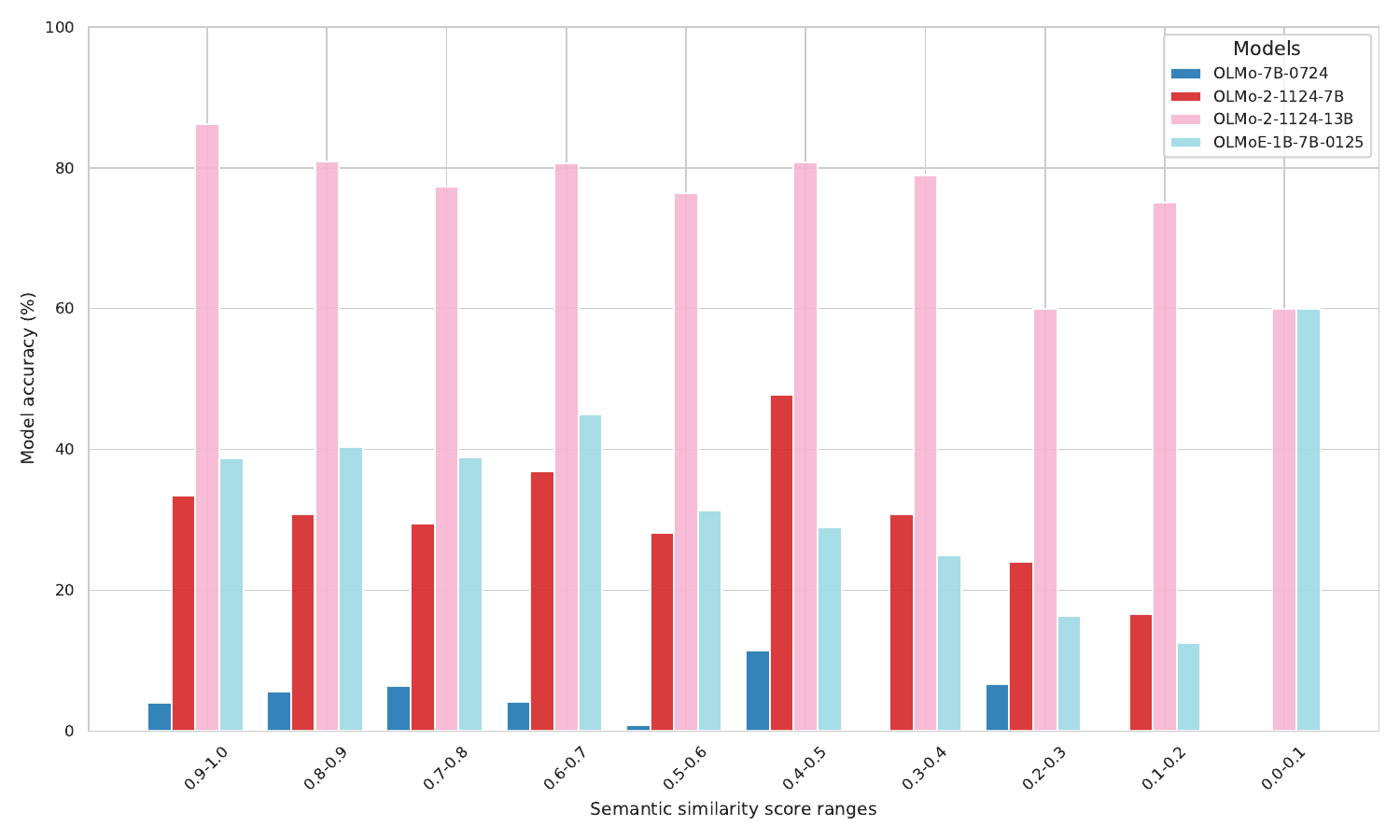}
        \caption{\textsc{BoolQ}}
    \end{subfigure}
    \hfill
    \begin{subfigure}[b]{0.3\textwidth}
        \includegraphics[width=\textwidth]{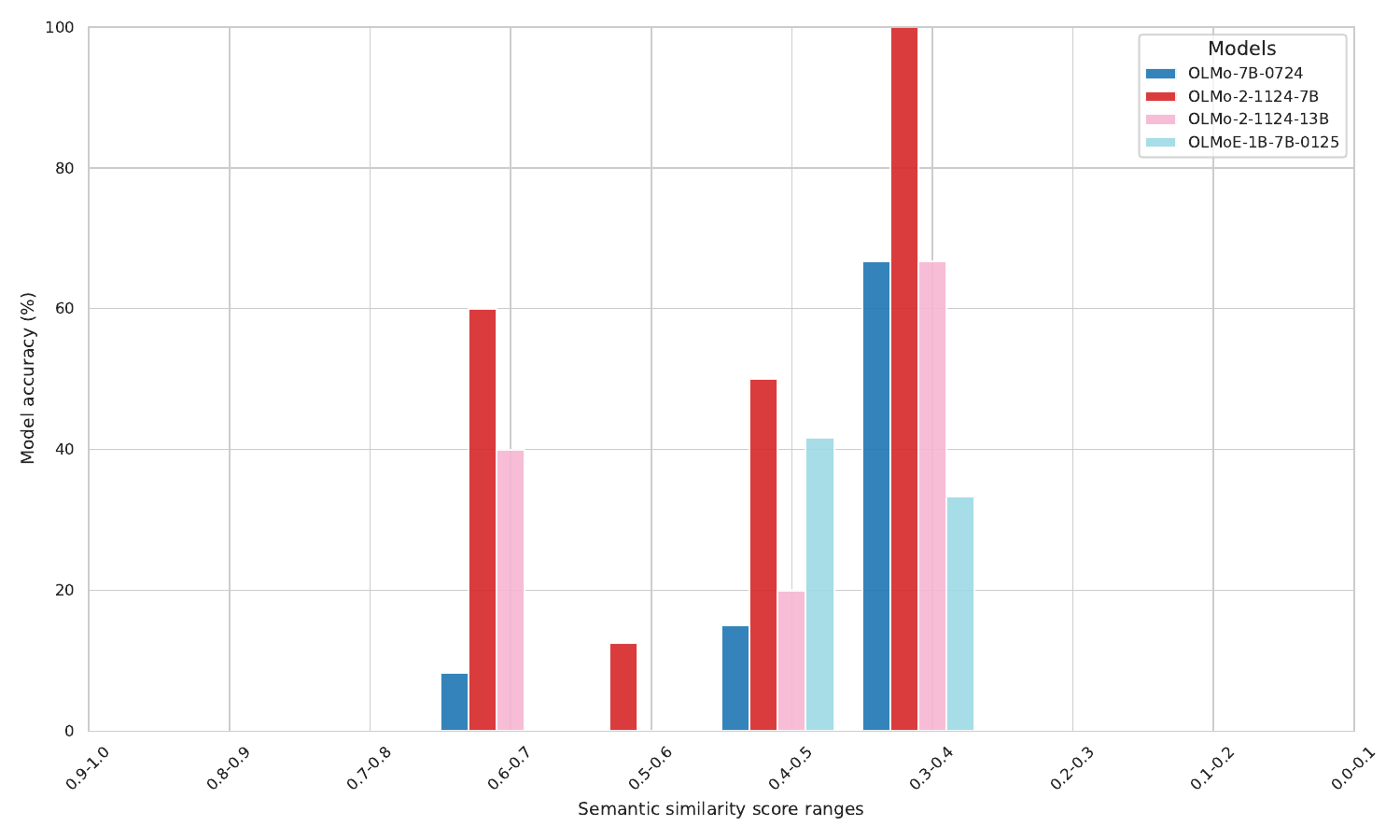}
        \caption{\textsc{WikiWhy}}
    \end{subfigure}
    \hfill
    \begin{subfigure}[b]{0.3\textwidth}
        \includegraphics[width=\textwidth]{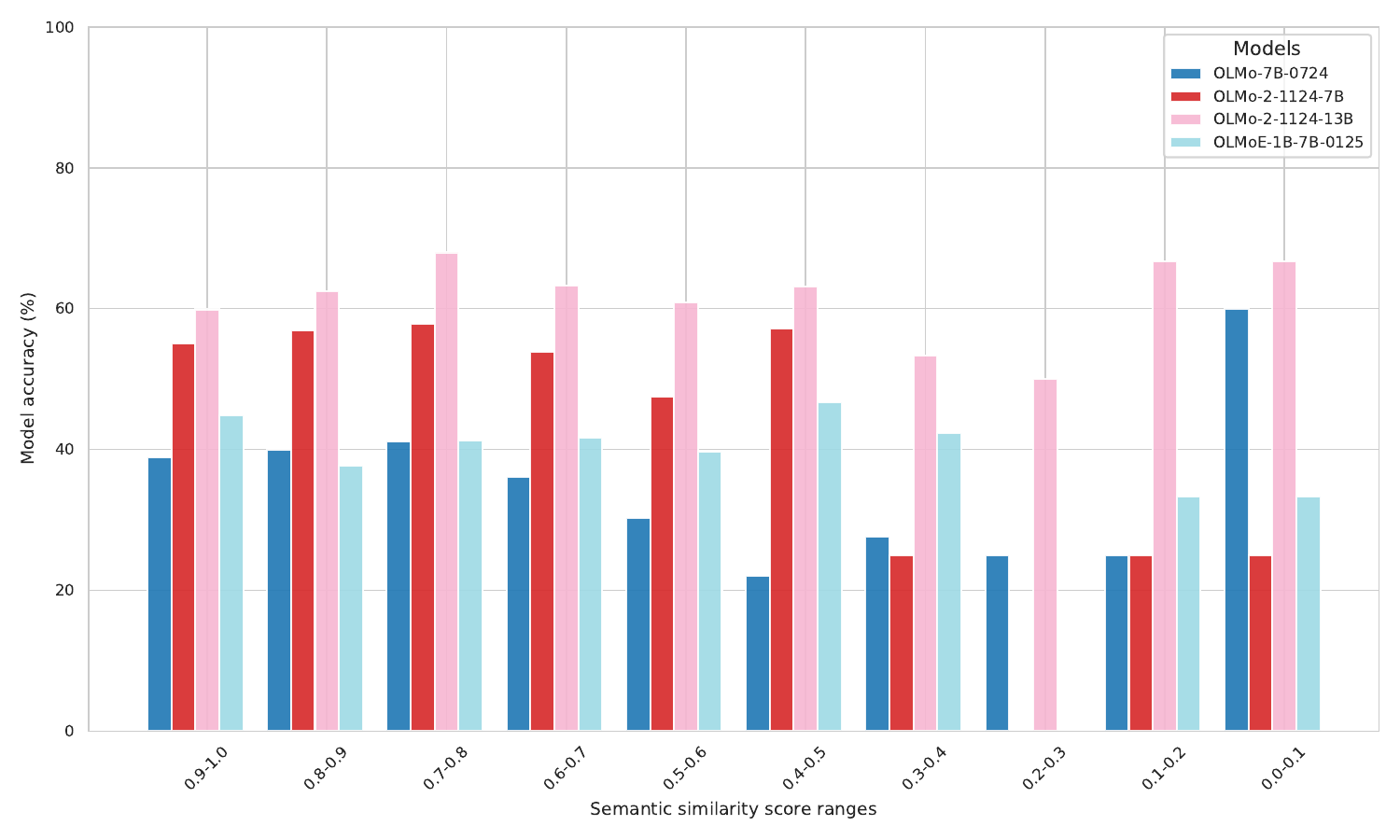}
        \caption{\textsc{HotpotQA}}
    \end{subfigure}

    \caption{Average accuracy of the Base \texttt{OLMo} LLMs across ten semantic similarity bins.}
    \label{fig:olmo_results_Base}
\end{figure*}

\begin{figure*}[htb!]
    \centering
    \begin{subfigure}[b]{0.3\textwidth}
        \includegraphics[width=\textwidth]{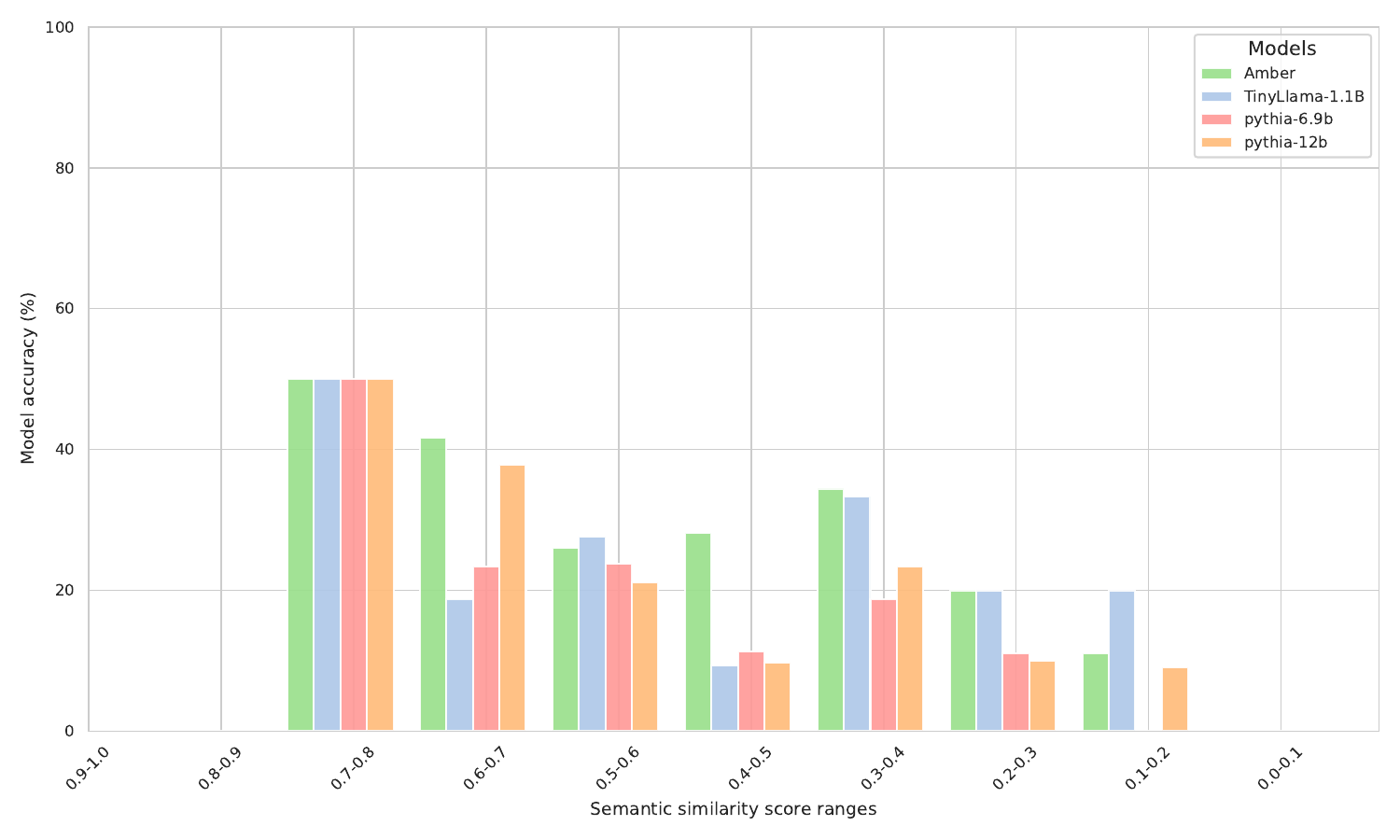}
        \caption{\textsc{SQuAD 2.0}}
    \end{subfigure}
    \hfill
    \begin{subfigure}[b]{0.3\textwidth}
        \includegraphics[width=\textwidth]{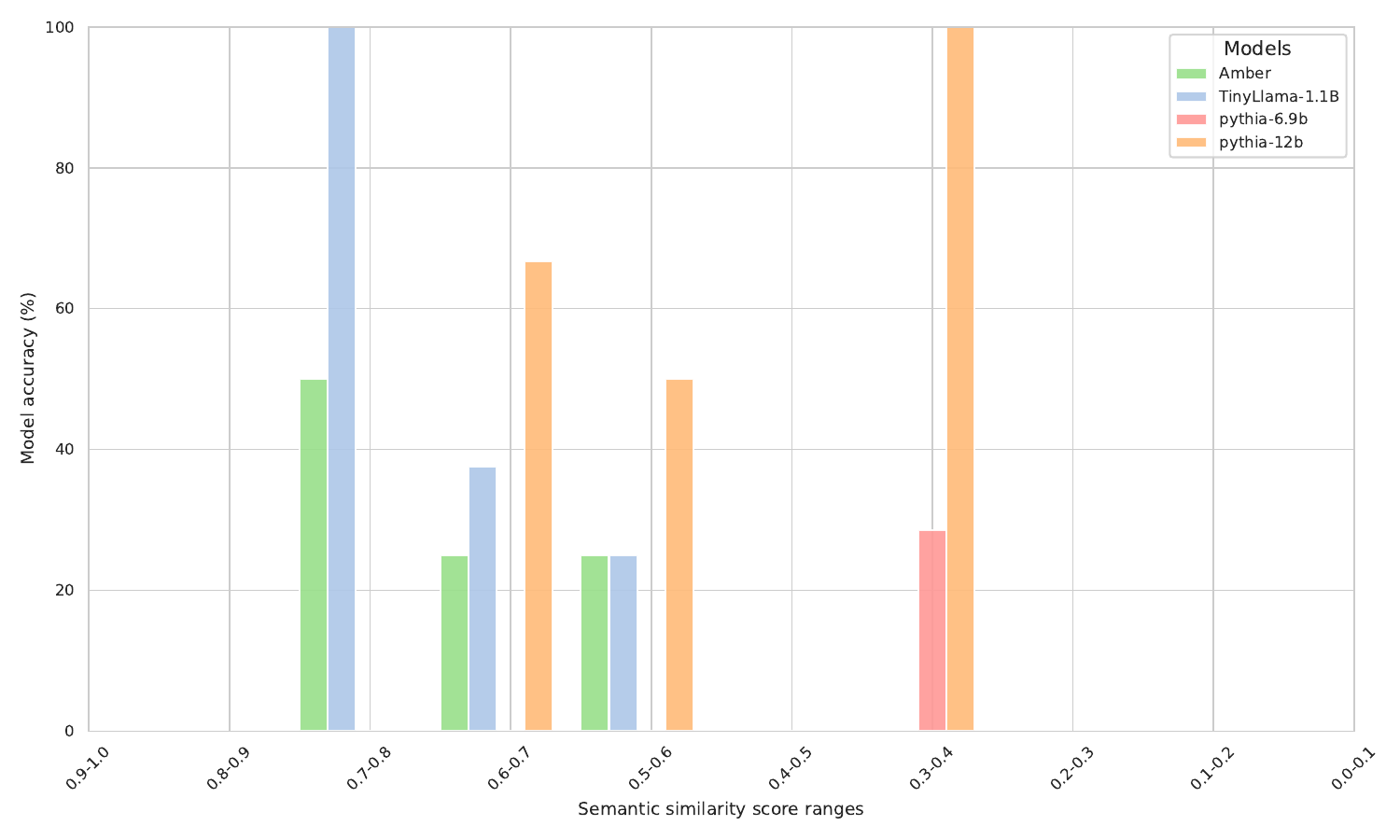}
        \caption{\textsc{WikiWhy}}
    \end{subfigure}
    \hfill
    \begin{subfigure}[b]{0.3\textwidth}
        \includegraphics[width=\textwidth]{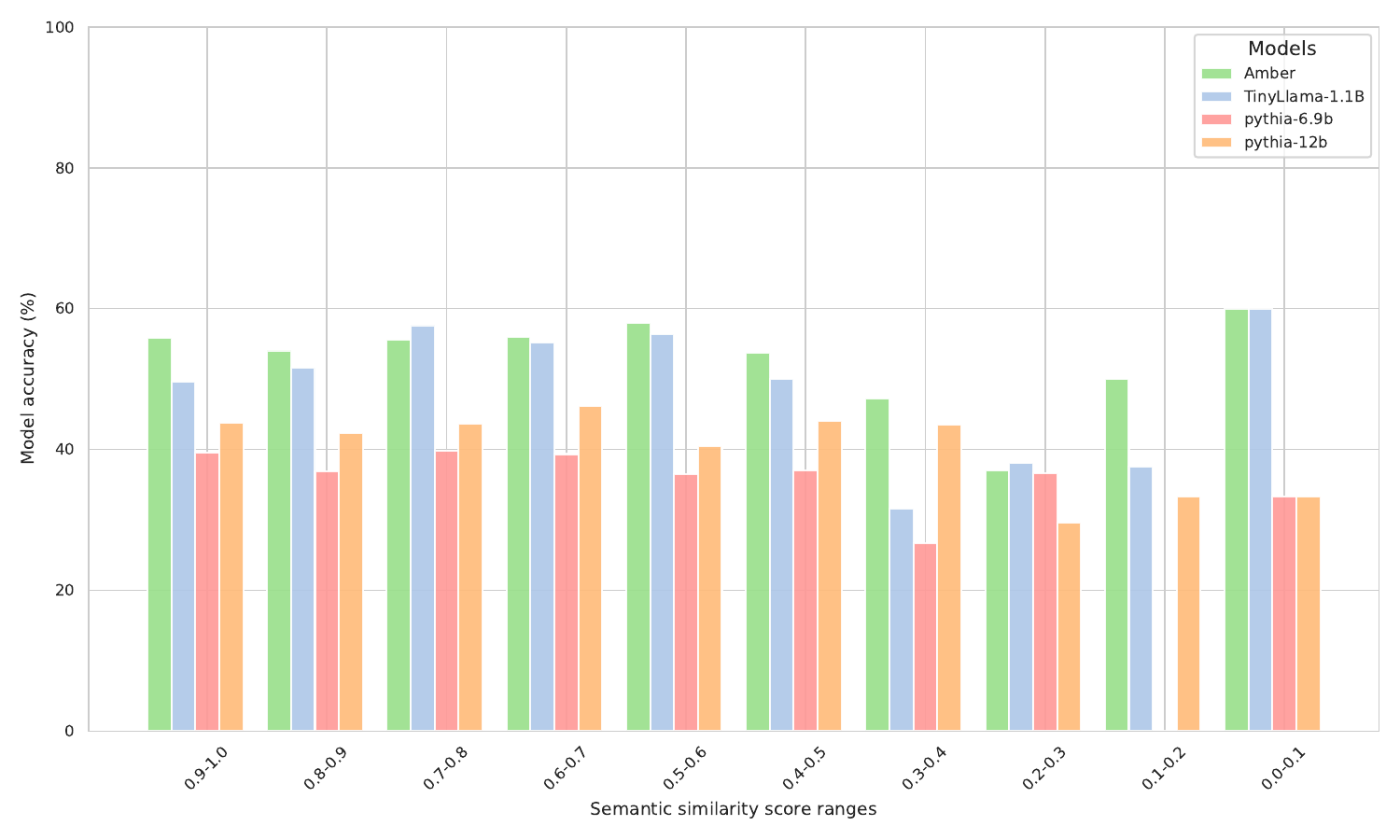}
        \caption{\textsc{HotpotQA}}
    \end{subfigure}
    \caption{Average accuracy of other Base LLMs across ten semantic similarity bins.}
    \label{fig:other_results_Base}
\end{figure*}

Figures~\ref{fig:olmo_results_Base} and~\ref{fig:other_results_Base} show that non-aligned base LLMs are also susceptible to the harmful effects of natural context drift.

\section{Chain-of-Thought Does Not Alleviate Drift-Induced Degradation}
\label{sec:Chain-of-Thought Does Not Alleviate Drift-Induced Degradation}

\begin{figure*}[htb!]
    \centering
    \begin{subfigure}[b]{0.3\textwidth}
        \includegraphics[width=\textwidth]{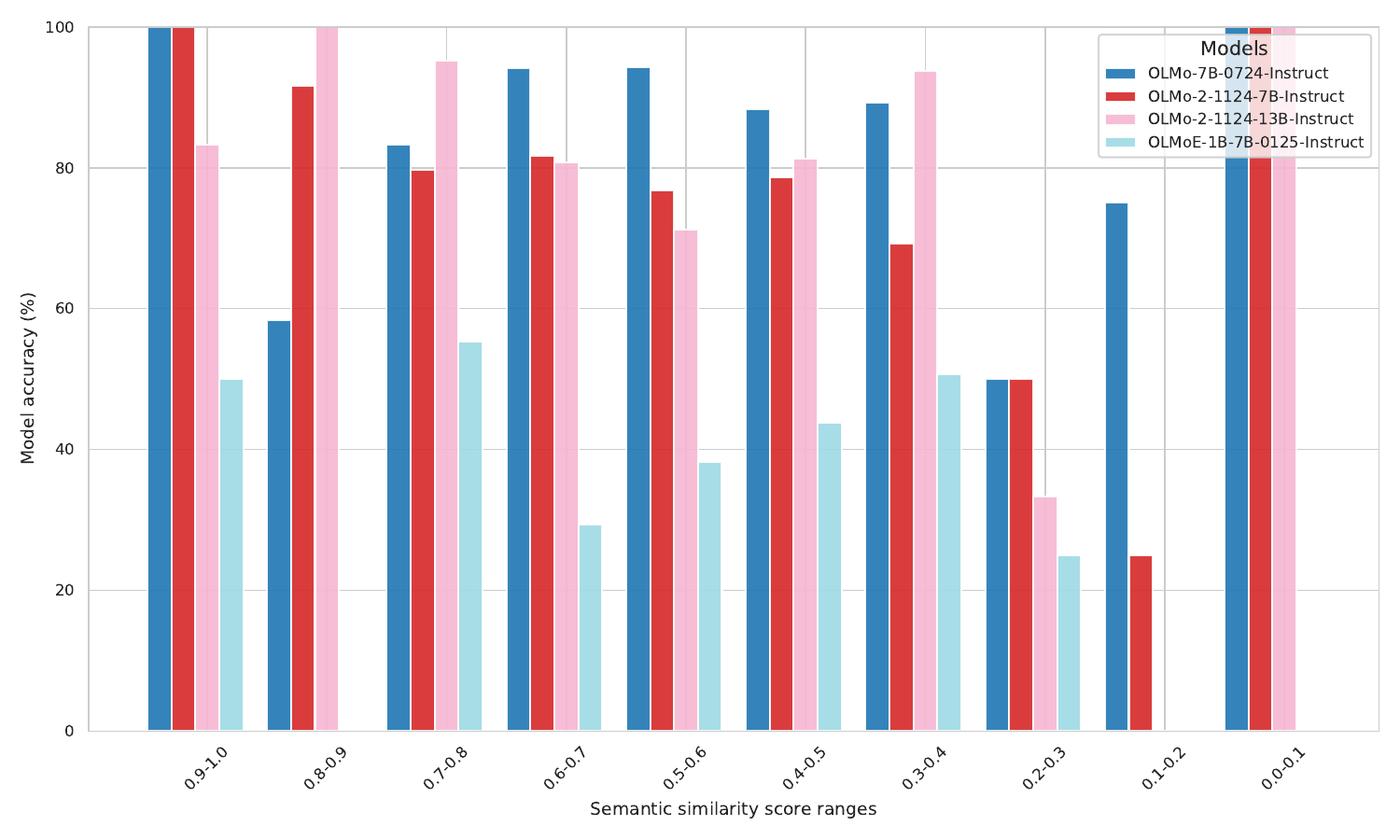}
        \caption{\textsc{SQuAD 1.1}}
    \end{subfigure}
    \hfill
    \begin{subfigure}[b]{0.3\textwidth}
        \includegraphics[width=\textwidth]{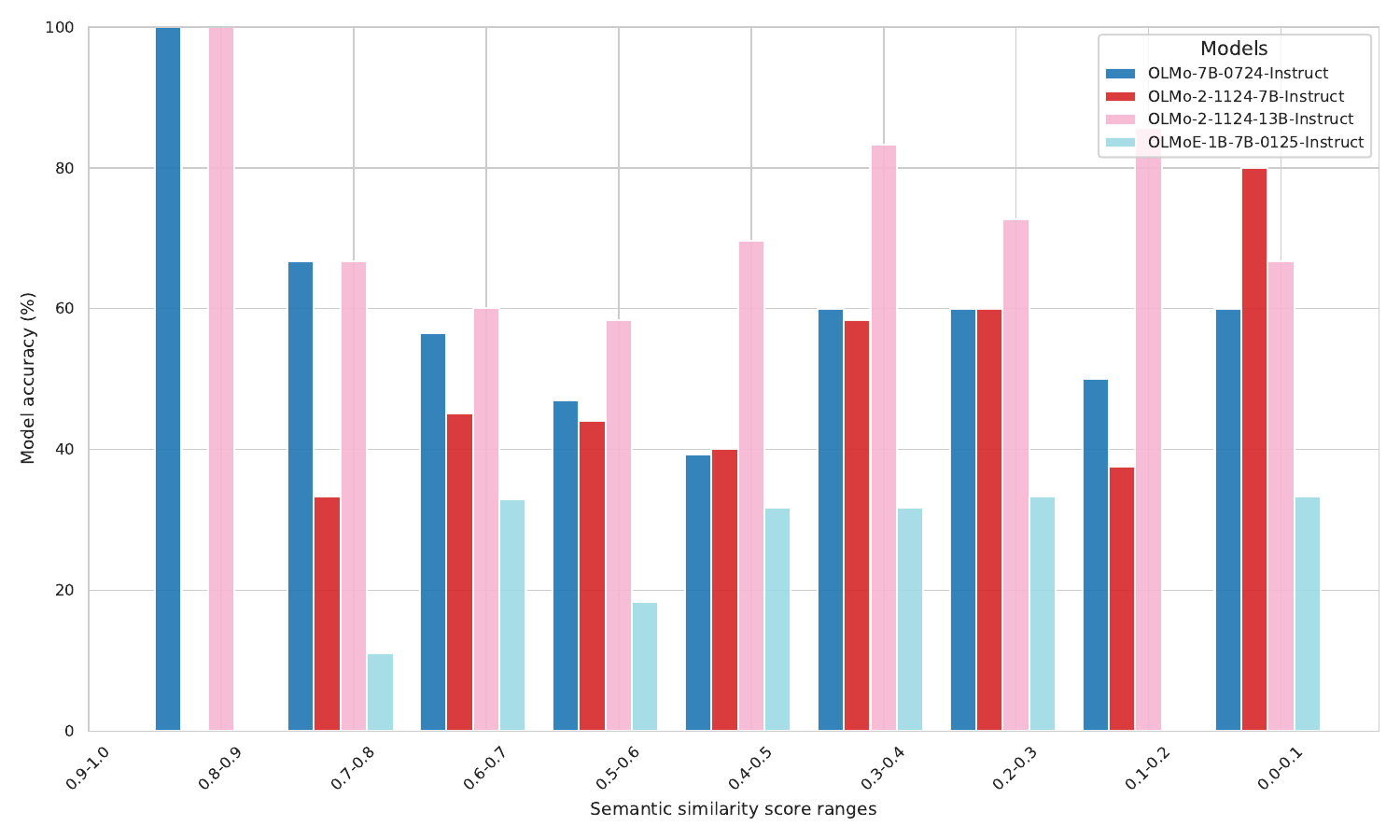}
        \caption{\textsc{SQuAD 2.0}}
    \end{subfigure}
    \hfill
    \begin{subfigure}[b]{0.3\textwidth}
        \includegraphics[width=\textwidth]{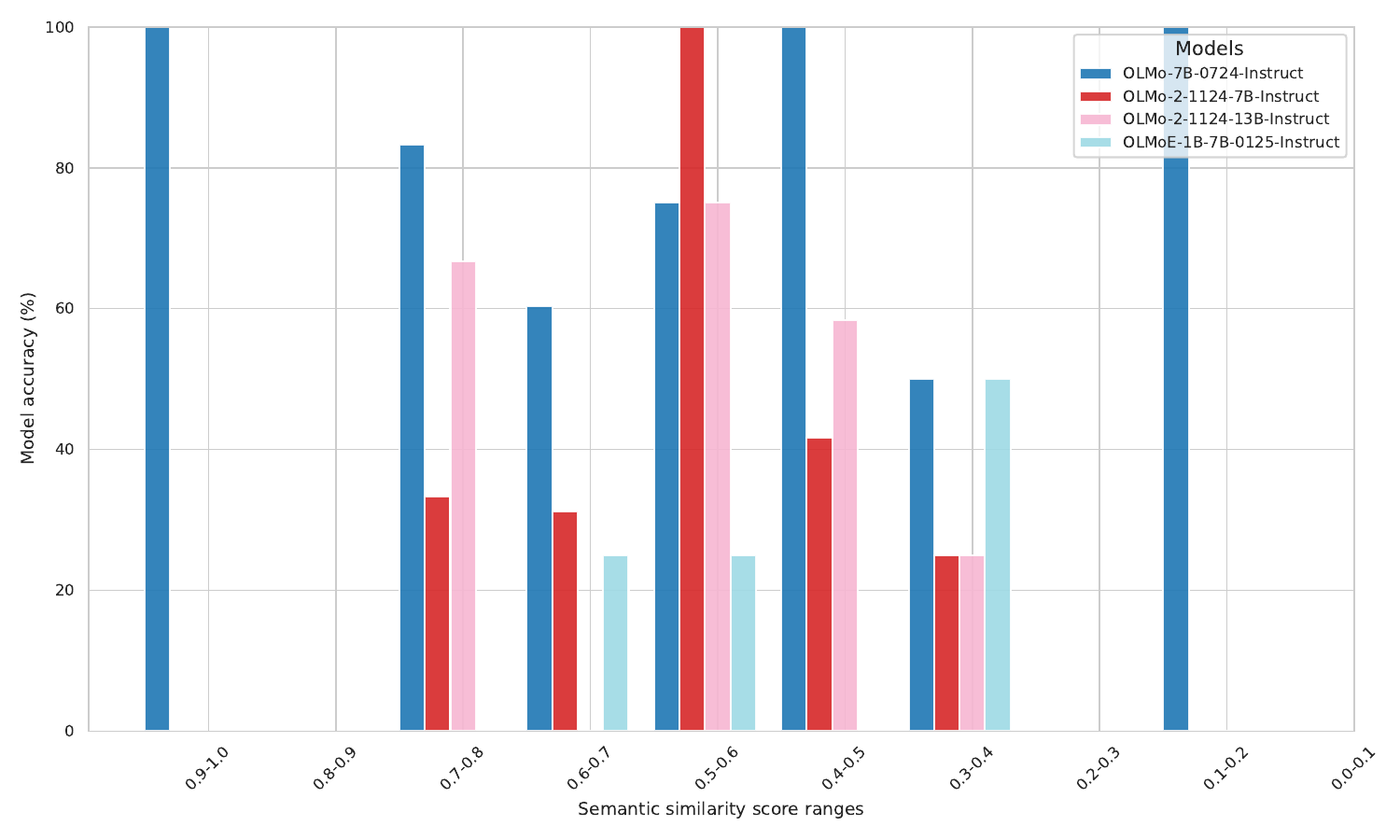}
        \caption{\textsc{D(RoBERTa)}}
    \end{subfigure}

    \vspace{0.5cm}

    \begin{subfigure}[b]{0.3\textwidth}
        \includegraphics[width=\textwidth]{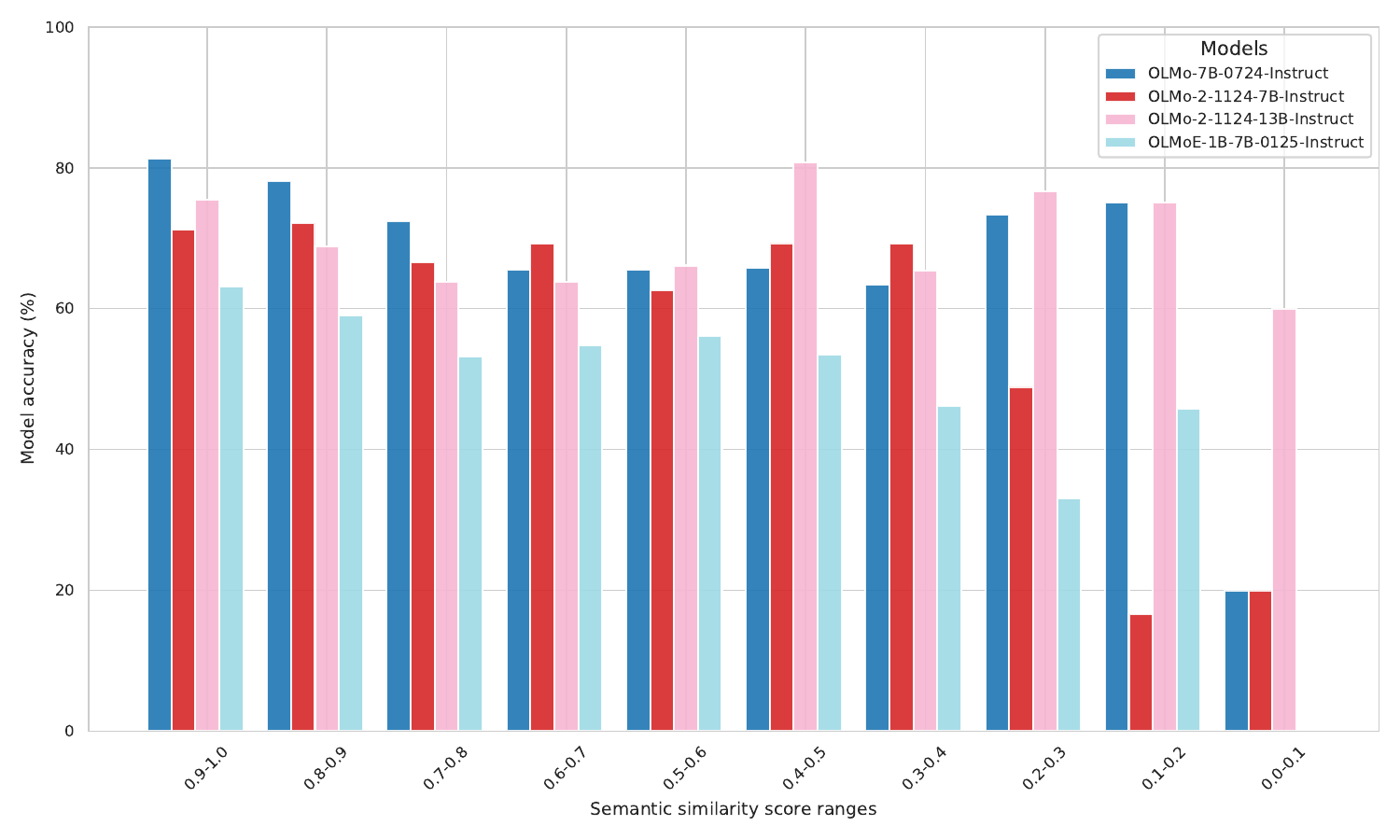}
        \caption{\textsc{BoolQ}}
    \end{subfigure}
    \hfill
    \begin{subfigure}[b]{0.3\textwidth}
        \includegraphics[width=\textwidth]{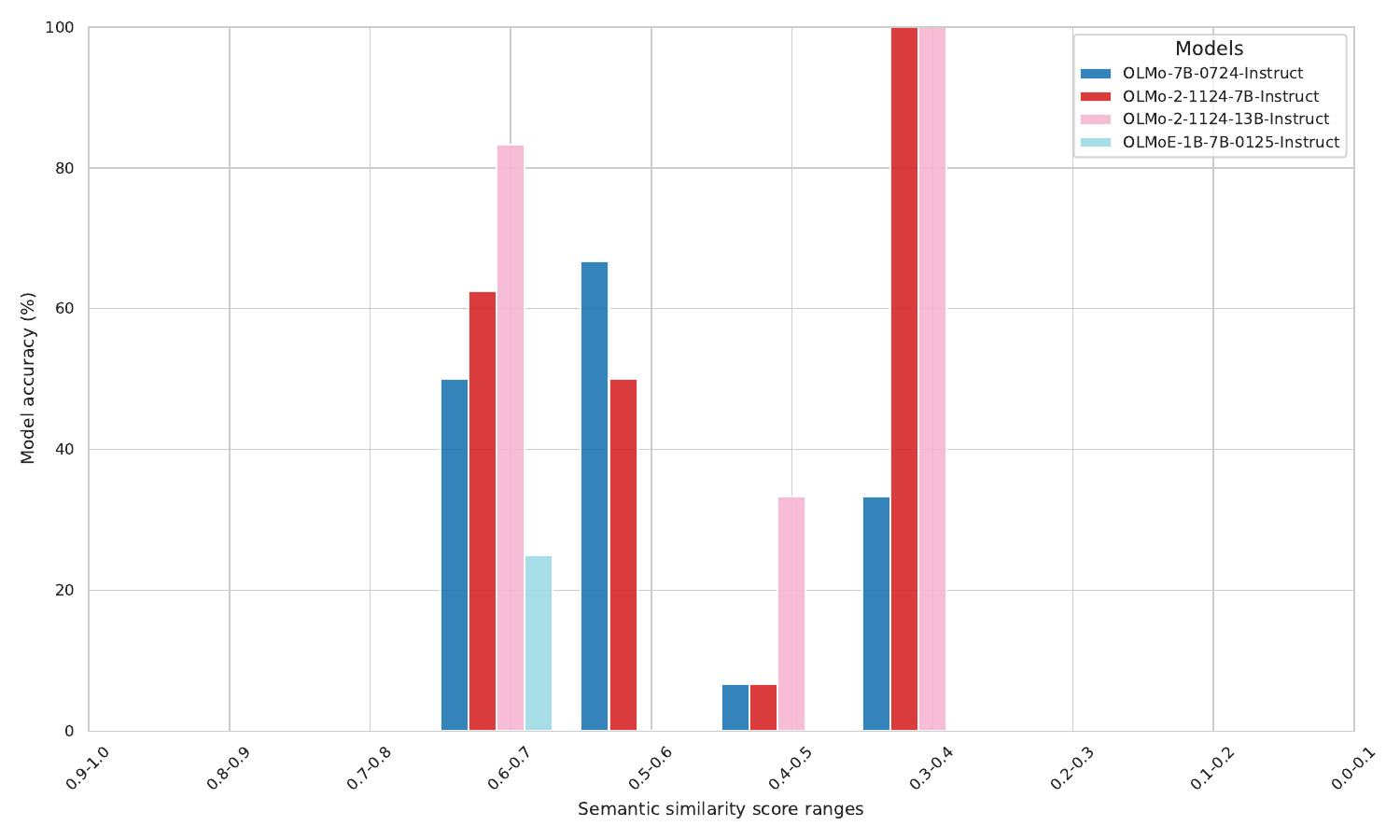}
        \caption{\textsc{WikiWhy}}
    \end{subfigure}
    \hfill
    \begin{subfigure}[b]{0.3\textwidth}
        \includegraphics[width=\textwidth]{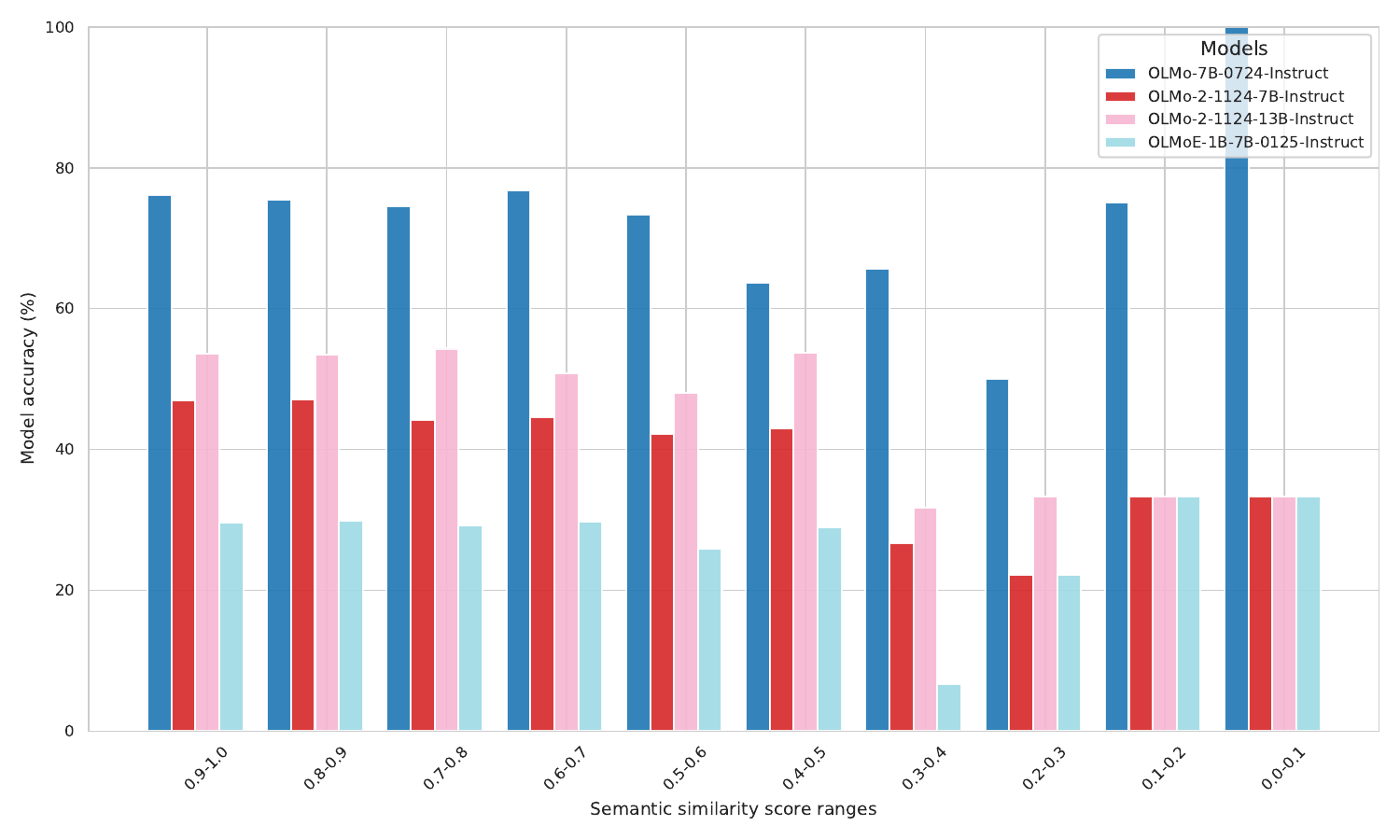}
        \caption{\textsc{HotpotQA}}
    \end{subfigure}

    \caption{Average accuracy of the instruction-finetuned \texttt{OLMo} LLMs across ten semantic similarity bins, with the prompt encouraging elaborated answers and CoT reasoning.}
    \label{fig:olmo_results_CoT}
\end{figure*}

\begin{figure*}[htb!]
    \centering
    \begin{subfigure}[b]{0.3\textwidth}
        \includegraphics[width=\textwidth]{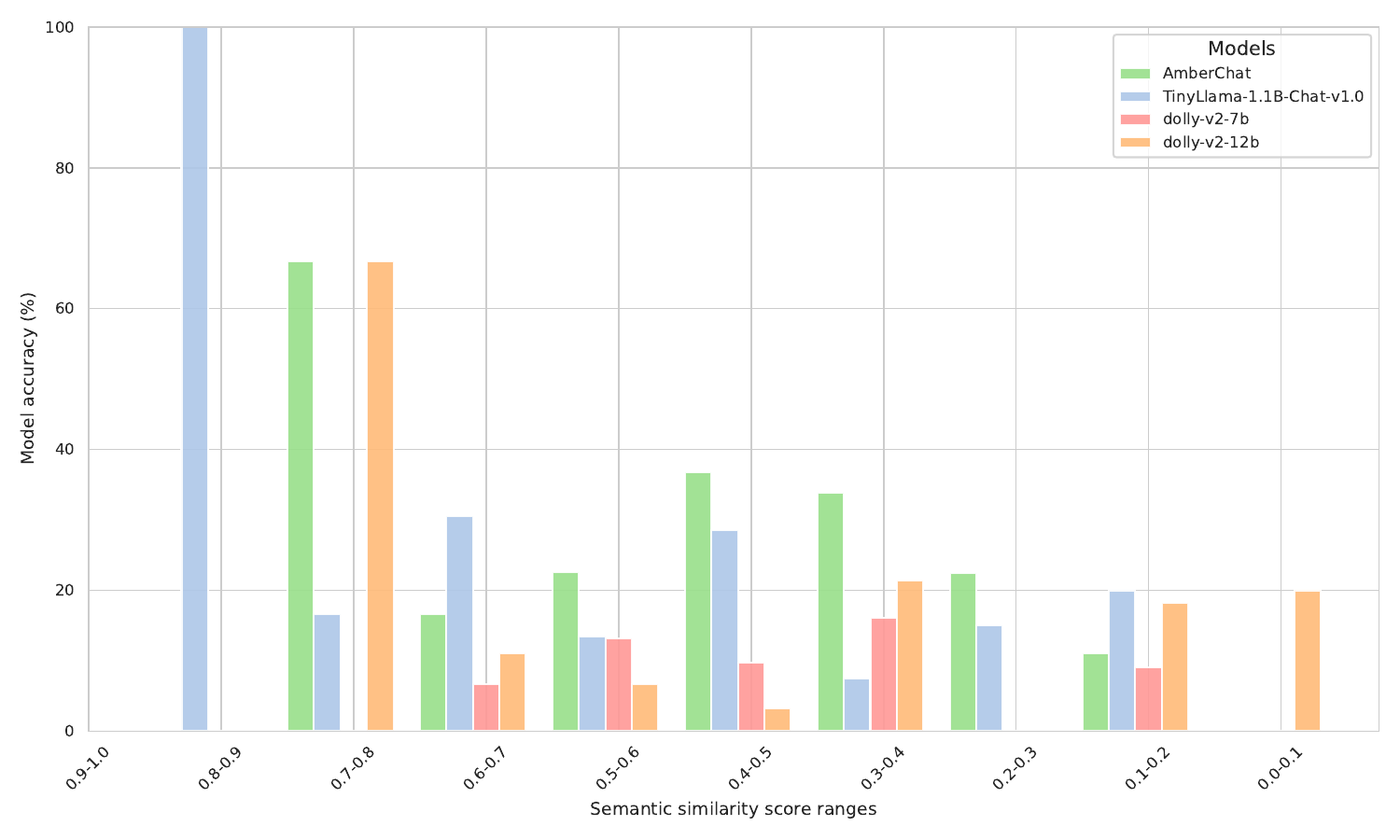}
        \caption{\textsc{SQuAD 2.0}}
    \end{subfigure}
    \hfill
    \begin{subfigure}[b]{0.3\textwidth}
        \includegraphics[width=\textwidth]{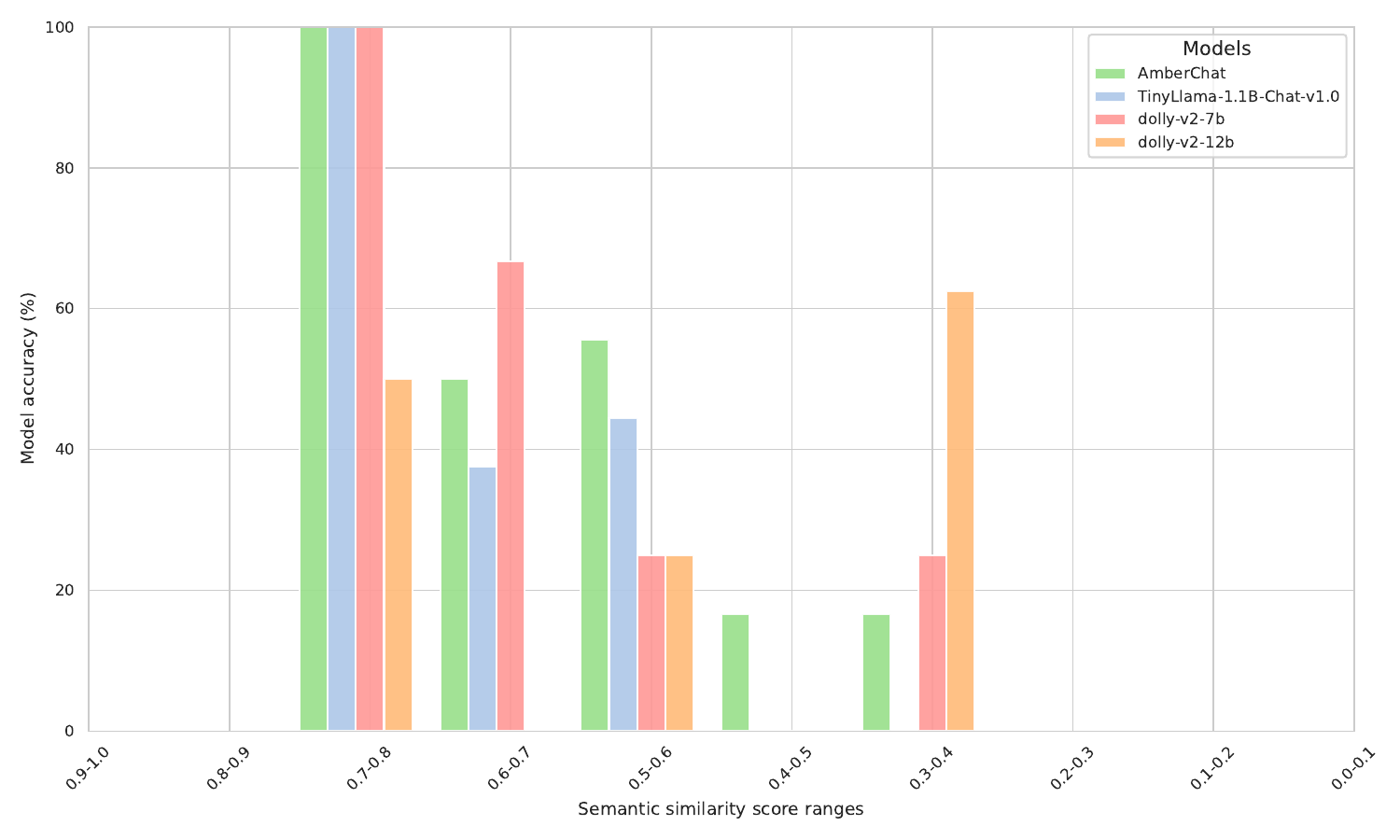}
        \caption{\textsc{WikiWhy}}
    \end{subfigure}
    \hfill
    \begin{subfigure}[b]{0.3\textwidth}
        \includegraphics[width=\textwidth]{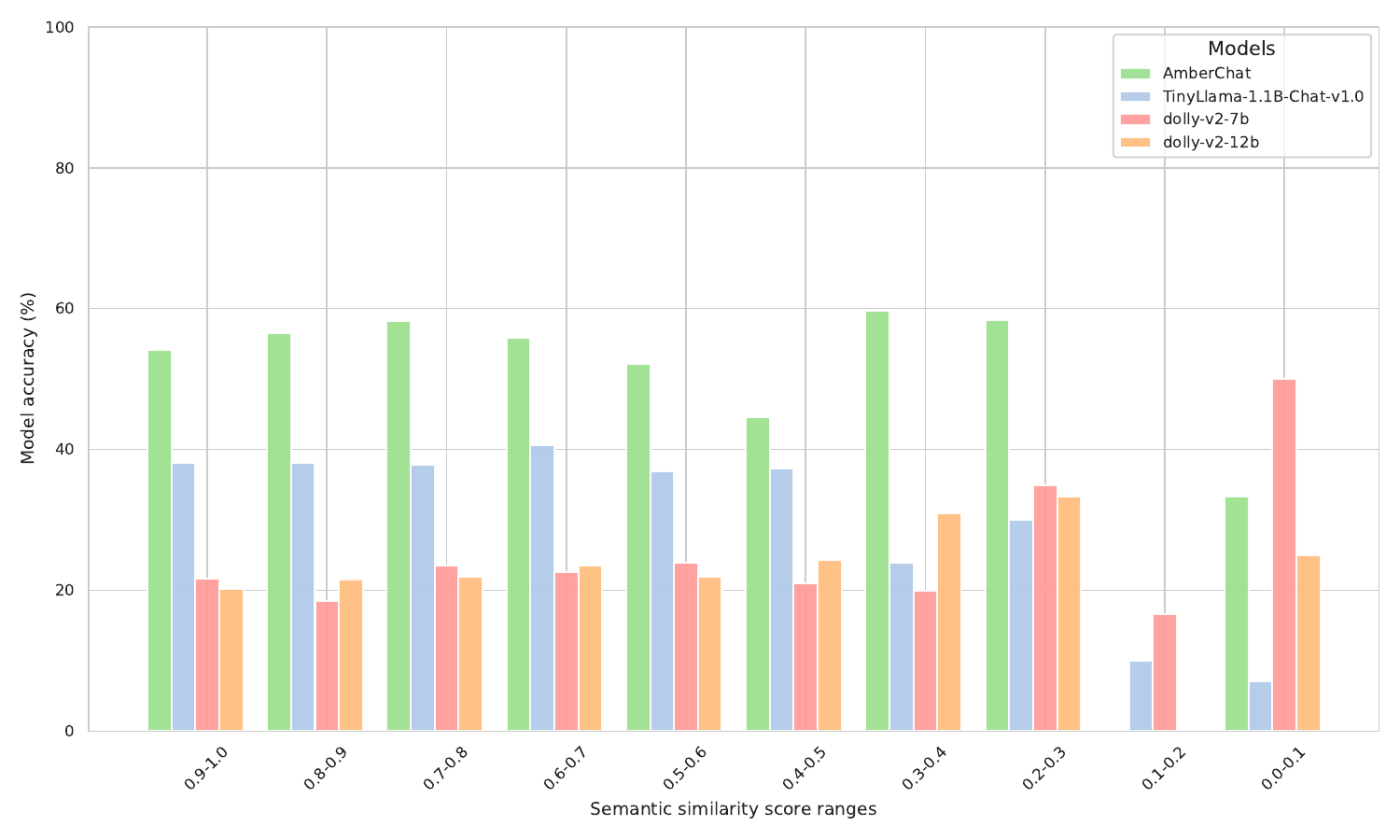}
        \caption{\textsc{HotpotQA}}
    \end{subfigure}
    \caption{Average accuracy of other instruction-finetuned LLMs across ten semantic similarity bins, with the prompt encouraging elaborated answers and CoT reasoning.}
    \label{fig:other_results_CoT}
\end{figure*}

\section{Human Annotation Details}
\label{sec:Human Annotation Details}

We adopt the same instructions provided to human annotators in \citep{wu2025payattentionrealworld} and recruit doctoral students from the university as annotators. All participants possess strong English reading comprehension skills. Before beginning the main annotation task, annotators are asked to label a small set of instances and resolve any disagreements through discussion until consensus is reached. During annotation, they are provided only with the edited reading paragraph and the corresponding question, without being informed that the paragraph has been modified, in order to minimize potential bias.

\end{document}